\documentclass[leqno,11pt]{article}
\usepackage{glottometrics}

\usepackage{hyperref}
\usepackage{color}
\usepackage{float}
\usepackage{tikz-dependency}
\depstyle{anderson}{draw=gray!80,rounded corners=11pt, inner sep=5pt, top color=white, bottom color=pink}
\usepackage{multicol}
\usepackage{lscape} 
\usepackage{colortbl}
\usepackage{multirow}

\usepackage{rotating}
\usepackage{etoolbox}
\newtoggle{Sonia} 
\togglefalse{Sonia}

\DeclareMathOperator*{\mean}{mean}

\DeclareMathAlphabet{\pazocal}{OMS}{zplm}{m}{n}

\glottovol{XX}
\glottoyear{20XX}
\glottodoi{Here will be added DOI number by Glottometrics.}

\runningauthors{Petrini \& Ferrer-i-Cancho}
\runningtitle{The distribution of syntactic dependency distances}

\title{The distribution of syntactic dependency distances}

\addauthor{1}{Sonia Petrini}{0000-0002-0514-6223}
\addauthor{1*}{Ramon Ferrer-i-Cancho}{0000-0002-7820-923X}

\affil{1}{Quantitative, Mathematical and Computational Linguistics Research Group. Departament de Ci\`encies de la Computaci\'o, Universitat Polit\`ecnica de Catalunya}

\correspond{*}{Corresponding author's email: rferrericancho@cs.upc.edu.}

\begin{document}
\maketitle


\begin{abstract}
The syntactic structure of a sentence can be represented as a graph, where vertices are words and edges indicate syntactic dependencies between them. In this setting, the distance between two linked words is defined as the difference between their positions. Here we wish to contribute to the characterization of the actual distribution of syntactic dependency distances, 
which has previously been argued to follow a power-law distribution.
Here we propose a new model with two exponential regimes in which the probability decay is allowed to change after a break-point. This transition could mirror the transition from the processing of word chunks to higher-level structures. We find that a two-regime model -- where the first regime follows either an exponential or a power-law decay -- is the most likely one in all 20 languages we considered, independently of sentence length and annotation style. Moreover, the break-point exhibits low variation across languages and averages values of 4-5 words, suggesting that the amount of words that can be simultaneously processed abstracts from the specific language to a high degree. The probability decay slows down after the breakpoint, consistently with a universal chunk-and-pass mechanism. Finally, we give an account of the relation between the best estimated model and the closeness of syntactic dependencies as function of sentence length, according to a recently introduced optimality score.
\end{abstract}






\section{Introduction}
\label{sec:Introduction}
Language is one of the most complex and fascinating expressions of humans as social animals, stemming from our urge for communication and physical and cognitive limitations. 
The interaction between these two forces inevitably shapes language 
at many levels \parencite{Christiansen2016a,Liu2017DependencyDA}. Among them we here focus on syntax, namely the way in which words in a sentence compose into larger hierarchical structures, creating a parallel dimension to their plain linear arrangement. The hierarchical structure arises from the relations between words
, modelled by means of a directed edge in the one-dimensional space of the network of a sentence (\autoref{fig:example_tree}). We call the resulting structure a syntactic dependency tree: each vertex is a word, and each word -- besides the root -- depends syntactically on its head, to which it is connected by an edge. We define $d$ as the absolute value of the difference between the positions of two syntactically related words \parencite{ref:Ferrer-i-Cancho2004}. Thus, consecutive words are at distance $1$, words separated by an intermediate word are at distance $2$ and so on.
For instance, in \autoref{fig:example_tree} ``John'' and ``gave'' are at distance 1, ``gave'' and ``painting'' are at distance 3, and so on.

\begin{figure}[H]
\centering
\begin{dependency}[theme = simple, group style =anderson]
    \begin{deptext}[column sep=.5cm,
    row sep=.1ex,nodes={draw=pink, shade, top color=pink, rounded corners}]
        John \& gave \& Bill \& the \& painting \& that \& Mary \& hated. \\
    \end{deptext}
    \depedge{2}{1}{1}
    \depedge{2}{3}{1}
    \depedge{2}{5}{3}
    \depedge{5}{4}{1}
    \depedge[edge end x offset=4pt]{8}{5}{3}
    \depedge{8}{6}{2}
    \depedge{8}{7}{1}
\end{dependency}
\caption{\label{fig:example_tree} Example of syntactic dependency tree. Edges are labelled with the value of the syntactic dependency distance between the words they connect.}
\end{figure}

A well-established principle of Dependency Distance minimization (DDm) has been consistently found in languages, implying the preference for short dependencies \parencite{ref:Ferrer-i-Cancho2004, Liu2008a, Futrell2015LargescaleEO, Omega}. 

\subsection{On the distribution of syntactic dependency distances}

The large body of evidence in favor of DDm suggests that there are universal patterns underlying language structure, 
which are likely to reflect the functioning of the human brain rather than features of specific languages. Here we focus on 
the probability distribution of syntactic dependency distances as a window to that functioning \parencite{Liu2017DependencyDA}.
Ferrer-i-Cancho 
described the probability of a syntactic dependency as an exponentially decaying function of distance for sentences of fixed length in Czech and Romanian \parencite{ref:Ferrer-i-Cancho2004,Ferrer2017d}. However, he made an interesting observation concerning a change in the speed of the decay: the probability of observing a dependency at distance 4-5 or more is higher than expected, in the sense that the decay slows down, which apparently contradicts the DDm principle itself. Later on, Liu proposed a power-law behaviour to describe the distribution of dependency distances in a Chinese treebank, considering sentences of mixed length \parencite{ref:HaitaoLiu} that was later refined as a modified power law with an additional parameter \parencite{Liu2009a}. A later cross-linguistic study covering 30 languages identified a power-law distribution for long sentences, and an exponential trend in short ones \parencite{LuHaitao:49}. These approaches illustrate the complexity of the analysed problem. Nevertheless, all these distributions have a similar shape, characterized by the dominance of very short distances and a long tail \parencite{Jiang2015TheEO}. The observed differences could hence derive from systematic discrepancies in sentence lengths, context, and annotation style, which all influence syntactic dependency distances \parencite{Jiang2015TheEO,Omega}. Moreover, power-laws can emerge from mixing other distributions, for instance from differently parameterized exponentials \parencite{Stumpf10022012}.
Hence the need -- expressed in various studies \parencite{ref:Ferrer-i-Cancho2004, riskofmixing, Jiang2015TheEO} -- to find the common ground of these results, analyzing the distribution of dependency distances while accounting for all these factors: considering both mixed and fixed sentence lengths in a large enough parallel corpus, while also controlling for annotation style. 

\subsection{Exponential distributions in nature}

An exponential distribution of syntactic dependency distances was predicted assuming a constraint on the average distance between syntactically related words that was justified in terms of cognitive economy \parencite{ref:Ferrer-i-Cancho2004}. At a lower cognitive level, the exponential distribution of projection distances between cortical areas has been justified in terms of a general principle of wiring  economy in neural networks \parencite{Ercsey2013a}.

It is worth framing our proposal of a two-regime exponential distribution for syntactic dependency distances in a broader setting where a breakpoint may indicate a boundary between local and non-local dynamics. A double exponential distribution for the average distance traversed by foraging ants is a robust phenomenon where the breakpoint separates risk-averse from risk-prone trajectories \parencite{Campos2016a}. A hypothesis for the origins of the breakpoint in the distribution of syntactic dependency distances is elaborated below.

\subsection{Short-term memory (STM) limitations}

Short-term memory (also called working memory), refers to a system, or a set of processes, holding mental representations temporarily available for
use in thought and action \parencite{Cowan2017a}.
G. Miller's classic article
set the grounds for research on a possible absolute constraint on the amount of information that can be temporarily stored in memory, and on the mechanisms enacted to cope with it \parencite{Miller1956}. 
The estimated values of this maximum span vary: $7 \pm 2$ \parencite{Miller1956}, $2-3$ \parencite{Lewis2005-LEWAAM-2} or $4 \pm 1$ \parencite{Cowan2001}. However, it is commonly argued that such variation reflects variation in the unit of measurement: Miller's $7 \pm 2$ \parencite{Miller1956} would correspond to the amount of information before being compressed while lower values would correspond to chunks or compressed information \parencite{Mathy2012a}.


These considerations on STM are particularly relevant in the scope of linguistic communication: communicating requires constantly receiving and processing new inputs, without losing reference to the previous ones. To illustrate this, suppose a left-to-right incremental processing of the sentence in \autoref{fig:example_tree}. Let an open dependency be one in which only one of the two elements that compose it has already appeared, and a closed dependency one in which both the head and the dependent have already been encountered. Then, in the context of dependency structure the success of communication depends on the ability to keep track of an open dependency while opening new ones, and without knowing {\em a priori} when it is going to be closed \parencite{Liu2017DependencyDA}. Notice that dependencies represent relations between words, which are necessary for the speaker to convey a complex message building it from smaller units (encoding), and for the listener to recover such message by understanding the 
subjacent structure
of the sequence of words (decoding). Thus, syntactic structure really reveals the way in which humans deal with physical limitations to be able to produce and process 
a potentially unbounded number of words. Christiansen and Chater provided an integrated framework to describe both the cognitive constraints affecting STM in language processing -- what they call the ``now-or-never bottleneck'' -- and the chunking strategy enacted to cope with them, which they refer to as ``chunk-and-pass'' mechanism \parencite{Christiansen2016a}. They collected a wide set of empirical results, describing the bottleneck as mainly arising from our short memory for auditory signals, the speed of new incoming linguistic input, and from memory limitations on sequence recalling tasks. According to the authors, to deal with these constraints the human cognitive system relies on a series of strategies. That is, as we receive new linguistic input, we eagerly process it by grouping units into chunks, and passing them at a more abstract level of representation; once a chunk has been integrated into the available knowledge hierarchy (\autoref{fig:example_tree}), a new one can be processed and again passed at higher representation levels. This model entails that chunking is required 
to store information for a longer time
while a single word would be an easily forgotten piece of de-contextualized information, grouping words together produces a meaningful abstract image, which can be related to the following incoming concept. This mechanism would thus guarantee effective and efficient communication, profoundly shaping the structure of language itself.

\subsection{Contribution}

The primary aim of this work is to test the hypothesis that dependency distances in languages are distributed following two exponential regimes, modelled by means of a two-regime geometric distribution, and that the break-point between the regimes is similar across languages. The proposal of two regimes is motivated both empirically and theoretically. On one hand, it builds on the observations by Ferrer-i-Cancho concerning a change in probabilistic decay \parencite{ref:Ferrer-i-Cancho2004}. On the other hand, the existence of two different regimes would be consistent with the widely accepted idea that words are being chunked in order to be processed \parencite{Christiansen2016a}. Indeed, in a commentary on the work by Christiansen \& Chater, 
Ferrer-i-Cancho had suggested a relation between his empirical observation and their processing framework, linking the chunking mechanism with the puzzling slowing down of probability decay in syntactic dependency distances after 4-5 words \parencite{Christiansen2016a}. 
Verifying this hypothesis opens the path for a deeper understanding of the distribution of syntactic dependency distances, and of how this could be influenced and shaped by universal constraints on memory.
Concerning the first point, we believe our work will contribute to the existing literature on the distribution of dependency distances, finding a common ground to previous results by accounting for the effect of sentence length, context, and annotation style \parencite{ref:Ferrer-i-Cancho2004, riskofmixing, Jiang2015TheEO}. In fact, we consider both the syntactic structure of sentences with a specific length, and of various sentence lengths jointly, performing the analysis on a parallel corpus following two alternative syntactic dependency annotation schemes.
The second point is related to one of the free parameters of our models, namely the break-point between the two regimes. If the change in probability is a mirror of the chunking mechanism enacted in language processing, the break-point we estimate could be a visible and direct statistical marker of the hypothesis advanced by \textcite{Christiansen2016a}. In particular, it may approximate the distance after which physical and cognitive limitations become too pressing, and the current chunk needs to be closed and encoded in memory, in order not to be overwritten by forthcoming information.
Therefore, looking at the homogeneity of the estimated break-point values across languages could shed light on general cognitive patterns. 
Formally, we aim to verify the following two-fold hypothesis
\begin{itemize}
    \item $H_1$. Syntactic dependency distances are distributed following two exponential regimes.
    \item $H_2$. The break-point between the two regimes  exhibits low variation across languages and within a language. 
\end{itemize}
Additionally, we further investigate the relation between the DDm principle and sentence length \parencite{ref:Ferrer-i-Cancho2021}, analysing how it is reflected in the shape of the distribution of syntactic dependency distances. We use $\Omega$, a recently introduced optimality score, to quantify the intensity of DDm \parencite{Omega}.

\subsection{Structure}

The remainder of the article is organized as follows.
In order to test $H_1$, we compare the fit of the proposed two-regime model against an ensemble of alternative distributions. 
Section \ref{sec:Models} presents the definitions of the models for the distribution of syntactic dependency distances. 
Section \ref{sec:Materials} provides a detailed description of the data while Section \ref{sec:Methods} details the methodology.
Section \ref{sec:Results} reports the results of the model selection on sentences of languages from distinct families and investigates the relation between the best model and the optimality of syntactic dependency distances. Finally, section \ref{sec:Discussion} discusses the findings, focusing on the verification of our hypotheses and on other general patterns while accounting for the observed cross-linguistic variability.  Section \ref{sec:Conclusions} summarises the major conclusions of this article.  

\section{Models}
\label{sec:Models}

\begin{figure}[H]
  \centering
    \includegraphics[width = 0.9\textwidth]{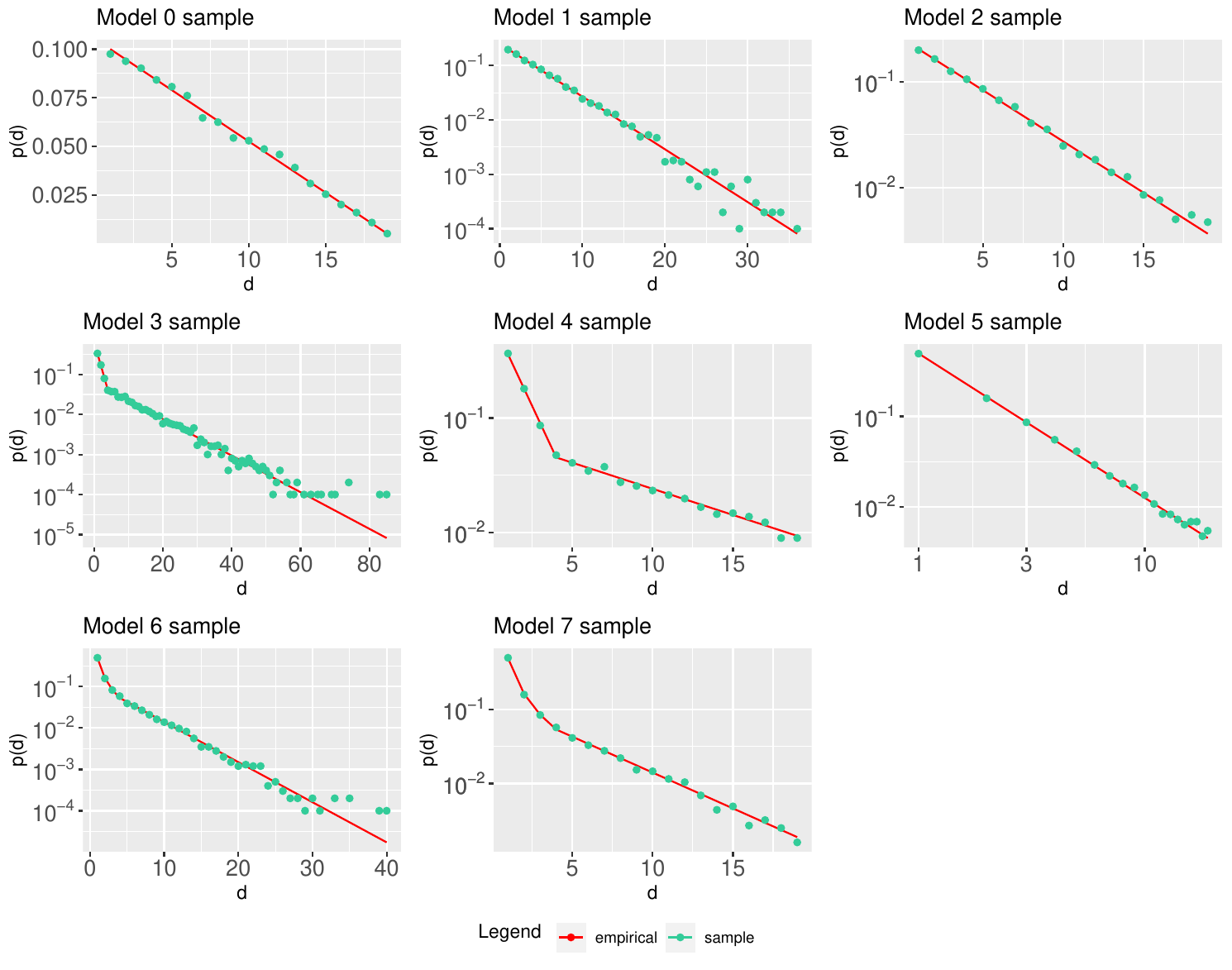}
    \caption{$p(d)$, the probability of $d$ in a model versus a random sample of itself. The random sample has size $10^4$. $n=20$ ($d_{max}=19$) for the right-truncated models. 
    \iftoggle{Sonia}{\textcolor{red}{Model 0 here could be Model 0 or Model 0.0}}{Thus Model 0 is the same as Model 0.0 here.}
    $d^*=4$ for the two-regime models. For the equations of the models refer to \autoref{tab:models}, while for the complete list of parameter values refer to  \autoref{tab:params_compare}.
    \iftoggle{Sonia}{\textcolor{red}{Remove "sample" from subfigure titles of the form "Model X samples".}}{}
    }
    \label{fig:artificial_data}
\end{figure}

We use $p(d)$ to refer to the probability that two linked words are at distance $d$. 
$d \in [1, n)$ in a sentence of $n$ words. 
See \autoref{tab:models} for a summary of the ensemble of models and \autoref{fig:artificial_data} for the shape of the models against an artificial random sample of their probability distributions (details on the generation of these samples are given in \autoref{sec:AppendixAB}).
Here we present a series of well-known models (e.g., geometric distribution, right-truncated zeta distribution) and non-standard models for $p(d)$. The details of the derivation of the non-standard models are given in \autoref{sec:AppendixA}.


\begin{table}[H]
\caption{\label{tab:models} Models for the distribution of syntactic dependency distances. $K$ is the number of free parameters. Refer to \autoref{sec:AppendixA} for the derivation of the equations.
}

\begin{tabular}{llcl} 
\toprule
\textbf{Model} & \textbf{Function} & $K$ & \textbf{Definition}  \\ 
\midrule
 0 & Null model  & 0 & $ \frac{1}{\binom{n}{2}}(n - d)$ if $d \in [1, n)$ \\
 0.0 & Null model  & 1 & $ \frac{1}{\binom{d_{max}+1} {2}}(d_{max}+1-d)$ if $d \in [1, d_{max}]$ \\  
 0.1 & Extended Null model & 0 & $\sum_{n=min(n)}^{max(n)} \frac{n-d}{\binom{n}{2}} \: p(n)$ if $d \in [1, max(n))$ \\  
 1 & Geometric  & 1 & $q(1 - q)^{d-1}$ if $d \geq 1$ \\  
 2 & Right-truncated geometric & 2  & $\frac{q(1 - q)^{d-1}}{1 - (1-q)^{d_{max}}}$ if $d \in [1, d_{max}]$\\  
 3 & Two-regime geometric & 3 & $\left\{
          \begin{array}{ll} 
              c_1(1 - q_1)^{d-1} & \mbox{if~} d \in [1, d_{max}] \\ 
              c_2(1 - q_2)^{d-1} & \mbox{if~} d \geq d^* \\ 
          \end{array}
        \right.$\\  
 4 & Two-regime - right-truncated geometric & 4 & $\left\{
          \begin{array}{ll} 
              c_1(1 - q_1)^{d-1} & \mbox{if~} d \in [1, d_{max}] \\ 
              c_2(1 - q_2)^{d-1} & \mbox{if~} d \in [d^*, d_{max}] \\ 
          \end{array}
        \right.$\\  
 5 & Right-truncated zeta distribution & 2  & $\frac{d^{-\gamma}}{H(d_{max}, \gamma)}$ if $d \geq 1$ \\  
 6 & Two-regime zeta-geometric & 3 & $\left\{
          \begin{array}{ll} 
              c_1 d^{-\gamma} & \mbox{if~} d \in [1, d_{max}] \\ 
              c_2(1 - q)^{d-1} & \mbox{if~} d \geq d^* \\ 
          \end{array}
        \right.$\\ 
 7 & Two-regime - right-truncated zeta-geometric & 4 & $\left\{
          \begin{array}{ll} 
              c_1 d^{-\gamma} & \mbox{if~} d \in [1, d_{max}] \\ 
              c_2(1 - q)^{d-1} & \mbox{if~} d \in [d^*, d_{max}] \\ 
          \end{array}
        \right.$\\ 
\bottomrule
\end{tabular}

\end{table}


The first model that we consider is Model 0, the null model obtained when a real sentence is shuffled at random or, equivalently, when there is no word order constraint (and all the $n!$ word orderings are equally likely). Then \parencite{ref:Ferrer-i-Cancho2004} 
\begin{equation}\label{eq:model0}
p(d) = \left\{ 
       \begin{array}{ll}
            \frac{1}{\binom{n}{2}}(n-d) & \mbox{if } d \in [1, n) \\
            0 & \mbox{otherwise}.
       \end{array}
       \right.
\end{equation}  
The formulation of Model 0 in \ref{eq:model0} assumes that the maximum distance is $n-1$ and that sentence length is unique, two assumptions that are too restrictive for our model selection setting. First, we do not know if actual maximum value of $d$ is $n - 1$ or a lower value that is unknown to us (but could be set by some memory limitations of the human brain).   
Second, we are interested in the best model by fixing sentence length (where sentence length is unique) and also when considering jointly all sentences of any length for a given language (where sentence length varies). 
Thus, for fitting purposes, we distinguish between two specifications of Model 0. In the first one, Model 0.0, we relax the first assumption and give the model the freedom to select a maximum distance that does not need to be $n - 1$, the theoretical maximum value of $d$. 
Accordingly, Model 0.0 is defined as
\begin{equation*}
p(d) = \left\{ 
       \begin{array}{ll}
            \frac{1}{\binom{d_{max}+1}{2}}(d_{max}+1-d) & \mbox{if } d \in [1, d_{max}] \\
            0 & \mbox{otherwise}
       \end{array}
       \right.
\end{equation*}
where $d_{max}$ is the only free parameter. 
The second specification of Model 0, Model 0.1 adapts the initial Model 0 to sentences of mixed lengths. Suppose that $p(n)$ is the proportion of sentences having length $n$, and $\min(n)$ and $max(n)$ are the minimum and maximum observed values of $n$ in the sample.
Then Model 0.1 is defined as 
\iftoggle{Sonia}{
\textcolor{red}{Sonia 1: please ensure that this is the actual formulation of the model as the previous formulae where not precise enough}}
\begin{equation*}
p(d) = \left\{ 
       \begin{array}{ll}
            \sum_{n=min(n)}^{max(n)} \frac{n - d}{\binom{n}{2}} \: p(n) & \mbox{if } d \in [1, max(n)) \\
            0 & \mbox{otherwise}.
       \end{array}
       \right.
\end{equation*}
The following models follow the same design principle of Model 0.0 and, for the sake of simplicity, do not introduce $n$ into the definition of the model as Model 0 or Model 0.1. 

\iftoggle{Sonia}{
\textcolor{red}{Sonia: revise the new more precise definitions of the models below}
}

Given that distances are discrete, an exponential decay can be modeled with a geometric curve. Thus, Model 1 is the displaced geometric distribution, defined as
\begin{equation}\label{eq:model1}
p(d) = \left\{ 
       \begin{array}{ll}
            q(1 - q)^{d-1} & \mbox{if } d \geq 1 \\
            0 & \mbox{otherwise},
       \end{array}
       \right. 
\end{equation}
where $q$ is the only free parameter. 
When $d \geq n$, the displaced geometric assumes that $p(d) > 0$ while in a real sentence $p(d) = 0$. For this reason, we also consider Model 2, that is a right-truncated version in which non-zero probability mass is restricted to $d \in [1, d_{max}]$, i.e. 
\begin{equation*}
p(d) = \left\{ 
       \begin{array}{ll}
            \frac{q(1 - q)^{d-1}}{1 - (1-q)^{d_{max}}} & \mbox{if } d \in [1, d_{max}) \\
            0 & \mbox{otherwise},
       \end{array}
       \right.
\end{equation*}
The two-regime models are obtained by splitting the range of variation of $d$ into two overlapping regimes, one for $1 \leq d \leq d^*$ and another for $d \geq d^*$, where  
$p'(d)$ and $p''(d)$, the probability mass in the first and in the second regime respectively, satisfy $p'(d^*) = p''(d^*)$. Accordingly,
Model 3 is a generalization of Model 1 that consists of two regimes, and is defined as 
\begin{equation*}
p(d) = \left\{
          \begin{array}{ll} 
              c_1(1 - q_1)^{d-1} & \mbox{if~} d \in [1, d^*] \\ 
              c_2(1 - q_2)^{d-1} & \mbox{if~} d \geq d^* \\ 
              0 & \mbox{otherwise}, 
          \end{array}
        \right.
\end{equation*}
where $c_1$ and $c_2$ are normalization factors defined as 
\begin{eqnarray}
c_1 & = & \frac{q_1q_2}{q_2 + (1-q_1)^{d^*-1}(q_1 - q_2) } \label{eq:c1_model3} \\
c_2 & = & \tau c_1 \nonumber \\
\tau & = & \frac{(1-q_1)^{d^*-1}}{(1-q_2)^{d^*-1}}. \label{eq:tau_models3-4}
\end{eqnarray}
Thus, the only free parameters of Model 3 are $q_1$, $q_2$ and $d^*$. 

Model 4 is a generalization of Model 3 by right truncation, that is 
\begin{equation*}
p(d) = \left\{
          \begin{array}{ll} 
              c_1(1 - q_1)^{d-1} & \mbox{if~} d \in [1, d^*]  \\ 
              c_2(1 - q_2)^{d-1} & \mbox{if~} d \in [d^*, d_{max}], \\ 
              0 & \mbox{otherwise}, 
          \end{array}
        \right.
\end{equation*}
where $c_1$ and $c_2$ are normalization factors defined as 
\begin{equation}\label{eq:c1_model4}
c_1 = \frac{q_1q_2}{q_2 + (1-q_1)^{d^*-1}(q_1 - q_2 - q_1(1-q_2)^{d_{max} - d^*+1} )}.
\end{equation}
and $c_2 = \tau c_1$ with $\tau$ defined as in \ref{eq:tau_models3-4}.
The only free parameters of Model 4 are $q_1$, $q_2$, $d^*$ and $d_{max}$.

Next, following previous on syntactic dependency distances \parencite{ref:HaitaoLiu}, we also consider Model 5, a power-law model that is a right-truncated zeta distribution with parameters $\gamma$ and $d_{max}$ \parencite{Wimmer1999}, that is defined as follows
\begin{equation*}
p(d) = \left\{
          \begin{array}{ll} 
              \frac{d^{-\gamma}}{H(d_{max}, \gamma)} & \mbox{if~} d \geq 1 \\ 
              0 & \mbox{otherwise},              
          \end{array}
        \right.
\end{equation*}
where 
\begin{equation*}
H(d_{max}, \gamma)=\sum _{k=1}^{d_{max}}{\frac {1}{k^{\gamma}}}
\end{equation*}
is the generalized harmonic number of order $\gamma$ of $d_{max}$.
Finally, we introduce Models 6 and 7, that are also composed of two regimes, the first one distributed as a right-truncated power-law and the second one as a geometric curve. Model 6 is defined as
\begin{equation*}
p(d) = \left\{
          \begin{array}{ll} 
              c_1 d^{-\gamma} & \mbox{if~} d \in [1, d^*] \\ 
              c_2(1 - q)^{d-1} & \mbox{if~} d \geq d^* \\ 
              0 & \mbox{otherwise},              
          \end{array}
        \right.
\end{equation*}
where $c_1$ and $c_2$ are normalization factors defined as 
\begin{eqnarray}
c_1 & = & \frac{q}{q H(d^*,\gamma) + d^*{^{-\gamma}} (1 - q)} \label{eq:c1_model6} \\ 
c_2 & = & \tau c_1 \nonumber \\
\tau & = & \frac{d^{*^{-\gamma}}}{(1-q)^{d^*-1}}. \label{eq:tau_models6-7} 
\end{eqnarray}

Model 7, the right-truncated version of Model 6, is defined as
\begin{equation*}
p(d) = \left\{
          \begin{array}{ll} 
              c_1 d^{-\gamma} & \mbox{if~} d \in [1, d^*] \\ 
              c_2(1 - q)^{d-1} & \mbox{if~} d \in [d^*, d_{max}] \\
              0 & \mbox{otherwise},
          \end{array}
        \right.
\end{equation*}
where 
\begin{equation}
\label{eq:c1_model7}
c_1 = \frac{q}{q H(d^*,\gamma) + 
d^{*^{-\gamma}} (1 - q - (1-q)^{d_{max} - d^* + 1} )},
\end{equation}
and $c_2 = \tau c_1$ with $\tau$ defined as in \ref{eq:tau_models6-7}.
The only free parameters of Model 6 are $\gamma$, $d^*$ and $q$. Model 7 adds a third free parameter that is $d_{max}$.

\subsection{Speed of decay}
\label{subsec:slope_q}
\begin{figure}[!htbp]
\centering
\includegraphics[width = 0.5\textwidth]{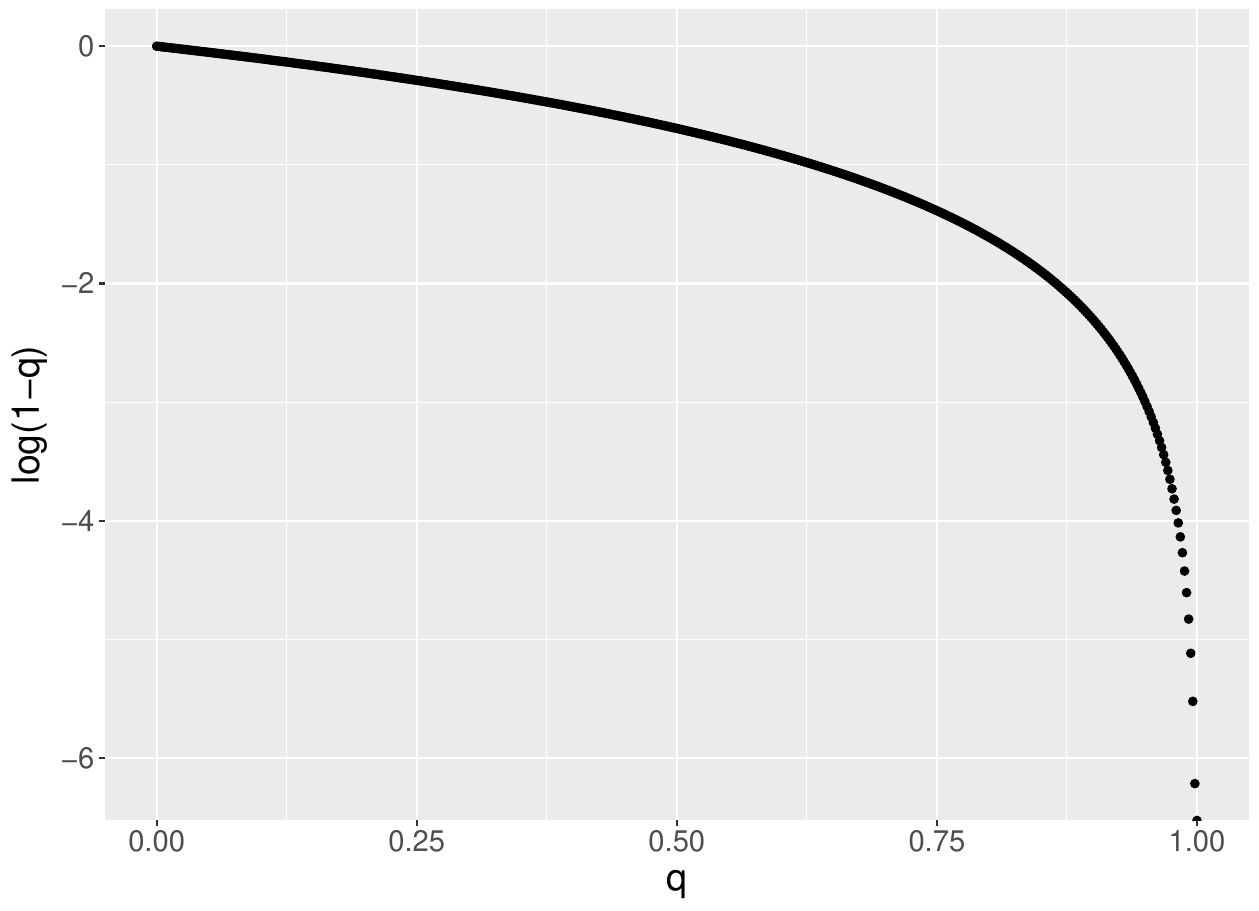}
\caption{\label{fig:q_plot} 
Slope of a geometric curve in log-linear scale as a function of its parameter $q$ 
\iftoggle{Sonia}{
\textcolor{red}{for $q \in [0, 1-\epsilon]$, with $\epsilon = ???$.} 
\textcolor{red}{Sonia 1: write the value of $\epsilon$ that you used to generate the plot to the previous equation} 
\textcolor{red}{Sonia 2: the current plot based on points suggest that the curve is discontinuous in a mathematical sense, which is not the case. Please, lines instead of points to actually show a continuous curve. Increase the sampling rate of the $[0,1]$ interval to improve the appearance of continuity if needed. }
\textcolor{red}{Sonia 3: somewhere it was written that the slope is $\log q(1-q)$. Please, ensure that the equation used to generate the figure is actually $\log (1-q)$. }
}
{for $q \in [0, 1)$.}
} 

\end{figure}

When plotted in log-linear scale, an exponential curve becomes a line. For a geometric model (\ref{eq:model1}), the slope of that line is $\log(1-q)$ since 
\begin{eqnarray*}
    \log p(d) & = & \log q(1-q)^{d-1} \\
    & = & d \log(1-q) + \log \frac{q}{1-q}.
\end{eqnarray*}
That slope conveys information about the speed of probability decay. Such slope is a decreasing function of $q$ (\autoref{fig:q_plot}), meaning that as $q$ increases the slope becomes more negative, and probability decays faster. In light of this fact, we consider parameters $q$ (Models 1 and 2) as well as $q_1$ and $q_2$ (Models 3-4) to account for the speed of exponential decay in the two regimes of Models 3-4, and we refer to them as ``slope parameters'' for simplicity. 

\section{Material}
\label{sec:Materials}

\begin{table}[H]
\caption{\label{tab:language_info} The languages, their linguistic family and their writing system. }
\centering
\begin{tabular}{@{}*{3}{l}}
\toprule
\textbf{Language} & \textbf{Family} & \textbf{Writing system}  \\ 
\midrule
 Arabic & Afro-Asiatic & Arabic \\ 
  Chinese & Sino-Tibetan & Han \\ 
  Czech & Indo-European & Latin \\ 
  English & Indo-European & Latin \\ 
  Finnish & Uralic & Latin \\ 
  French & Indo-European & Latin \\ 
  German & Indo-European & Latin \\ 
  Hindi & Indo-European & Devanagari \\ 
  Icelandic & Indo-European & Latin \\ 
  Indonesian & Austronesian & Latin \\ 
  Italian & Indo-European & Latin \\ 
  Japanese & Japonic & Japanese \\ 
  Korean & Koreanic & Hangul \\ 
  Polish & Indo-European & Latin \\ 
  Portuguese & Indo-European & Latin \\ 
  Russian & Indo-European & Cyrillic \\ 
  Spanish & Indo-European & Latin \\ 
  Swedish & Indo-European & Latin \\ 
  Thai & Kra-Dai & Thai \\ 
  Turkish & Turkic & Latin \\ 
\bottomrule
\end{tabular}
\end{table}

We extract syntactic dependency distances from a parallel subset of 20 languages from the Universal Dependencies collection \parencite{nivre-etal-2017-universal}. See \autoref{tab:language_info} for the languages, their linguistic family and their writing system. This subset is parallel in the sense that it contains the same sentences translated into every language. We use version 2.6, available \href{https://cqllab.upc.edu/lal/universal-dependencies/}{here}. Parallelism is crucial for robust cross-linguistic comparisons, as context can largely influence various aspects of language, including dependency structure. Another factor that shall be considered is annotation style, as there is no univocal way to generate syntactic dependency trees starting from a sentence. For this reason, we compare two different annotation styles: Universal Dependencies \parencite{nivre-etal-2017-universal} and the alternative Surface Syntactic Universal Dependencies \parencite{ref:SUD}. We refer to the two resulting versions of the collection as PUD and PSUD. See \autoref{tab:PUD_summary} and \autoref{tab:PSUD_summary} for a summary of the main statistical features of PUD and PSUD respectively. It can be seen that mean dependency distance values 
($\mean(d)$)
are smaller in PSUD.

 
\begin{table}[H]
\caption{\label{tab:PUD_summary} Summary of PUD collection. $\#s$ stands for number of sentences, $\#d$ stands for number of distances.}
\centering
\begin{tabular}{lrrrrrrrr}
\toprule
\textbf{Language}  & \textbf{$\# s$} & \textbf{$\# d$} & \textbf{$\min(d)$} & $\mean(d)$ & \textbf{$\max(d)$} & \textbf{$\min(n)$} & $\mean(n)$ & \textbf{$\max(n)$} \\ 
\midrule
 Arabic & 995 & 17514 &   1 & 2.30 &  30 &   3 & 18.60 &  50 \\ 
  Czech & 995 & 14976 &   1 & 2.39 &  29 &   3 & 16.05 &  44 \\ 
  German & 995 & 17544 &   1 & 3.11 &  42 &   4 & 18.63 &  50 \\ 
  English & 995 & 17711 &   1 & 2.53 &  31 &   4 & 18.80 &  56 \\ 
  Finnish & 995 & 12465 &   1 & 2.24 &  21 &   3 & 13.53 &  39 \\ 
  French & 995 & 21165 &   1 & 2.52 &  36 &   4 & 22.27 &  54 \\ 
  Hindi & 995 & 20517 &   1 & 3.30 &  42 &   4 & 21.62 &  58 \\ 
  Indonesian &995 & 16311 &   1 & 2.26 &  27 &   3 & 17.39 &  47 \\ 
  Icelandic & 995 & 15860 &   1 & 2.32 &  34 &   3 & 16.94 &  52 \\ 
  Italian & 995 & 20413 &   1 & 2.48 &  35 &   3 & 21.52 &  60 \\ 
  Japanese & 995 & 24703 &   1 & 2.97 &  65 &   4 & 25.83 &  70 \\ 
  Korean & 995 & 13978 &   1 & 2.75 &  37 &   3 & 15.05 &  43 \\ 
  Polish & 995 & 14720 &   1 & 2.23 &  27 &   3 & 15.79 &  39 \\ 
  Portuguese & 995 & 19808 &   1 & 2.53 &  34 &   4 & 20.91 &  58 \\ 
  Russian & 995 & 15369 &   1 & 2.27 &  32 &   3 & 16.45 &  47 \\ 
  Spanish & 995 & 19986 &   1 & 2.50 &  32 &   3 & 21.09 &  58 \\ 
  Swedish & 995 & 16119 &   1 & 2.47 &  31 &   4 & 17.20 &  49 \\ 
  Thai& 995 & 21034 &   1 & 2.44 &  38 &   4 & 22.14 &  63 \\ 
  Turkish & 995 & 13727 &   1 & 2.91 &  34 &   3 & 14.80 &  37 \\ 
  Chinese & 995 & 17501 &   1 & 3.09 &  39 &   3 & 18.59 &  49 \\ 
\bottomrule
\end{tabular}
\end{table}

 
 \begin{table}[H]
\caption{\label{tab:PSUD_summary} Summary of PSUD collection. $\#s$ stands for number of sentences, $\#d$ stands for number of distances.}
\centering
\begin{tabular}{lrrrrrrrr} 
\toprule
    \textbf{Language} & \textbf{$\#s$} & \textbf{$\#d$} & \textbf{$\min(d)$} & $\mean(d)$ & \textbf{$\max(d)$} & \textbf{$\min(n)$} & $\mean(n)$ & \textbf{$\max(n)$} \\ 
\midrule
 Arabic & 995 & 17514 &   1 & 2.05 &  30 &   3 & 18.60 &  50 \\ 
  Czech & 995 & 14976 &   1 & 2.11 &  29 &   3 & 16.05 &  44 \\ 
  German & 995 & 17544 &   1 & 2.82 &  38 &   4 & 18.63 &  50 \\ 
  English & 995 & 17711 &   1 & 2.12 &  31 &   4 & 18.80 &  56 \\ 
  Finnish & 995 & 12465 &   1 & 2.04 &  22 &   3 & 13.53 &  39 \\ 
  French &  995 & 21165 &   1 & 2.13 &  35 &   4 & 22.27 &  54 \\ 
  Hindi &  995 & 20517 &   1 & 3.04 &  38 &   4 & 21.62 &  58 \\ 
  Indonesian & 995 & 16311 &   1 & 2.00 &  27 &   3 & 17.39 &  47 \\ 
  Icelandic &  995 & 15860 &   1 & 1.92 &  34 &   3 & 16.94 &  52 \\ 
  Italian &  995 & 20413 &   1 & 2.10 &  35 &   3 & 21.52 &  60 \\ 
  Japanese & 995 & 24703 &   1 & 2.73 &  67 &   4 & 25.83 &  70 \\ 
  Korean & 995 & 13978 &   1 & 2.70 &  38 &   3 & 15.05 &  43 \\ 
  Polish & 995 & 14720 &   1 & 2.00 &  27 &   3 & 15.79 &  39 \\ 
  Portuguese &  995 & 19808 &   1 & 2.13 &  34 &   4 & 20.91 &  58 \\ 
  Russian &  995 & 15369 &   1 & 2.05 &  32 &   3 & 16.45 &  47 \\ 
  Spanish &  995 & 19986 &   1 & 2.13 &  31 &   3 & 21.09 &  58 \\ 
  Swedish & 995 & 16119 &   1 & 2.07 &  31 &   4 & 17.20 &  49 \\ 
  Thai &  995 & 21034 &   1 & 2.20 &  39 &   4 & 22.14 &  63 \\ 
  Turkish & 995 & 13727 &   1 & 2.86 &  33 &   3 & 14.80 &  37 \\ 
  Chinese & 995 & 17501 &   1 & 2.99 &  39 &   3 & 18.59 &  49 \\ 

\bottomrule
    \end{tabular}
\end{table}


\section{Methodology}
\label{sec:Methods}

The code for this work was written both in \texttt{R} and \texttt{python}, and is available \href{https://github.com/soniapetrini/DistributionOfDependencyDistances}{here}.

\subsection{Model selection}\label{subsec:model_selection}

We here describe the model selection procedure implemented to test $H_1$. This methodology is validated with the help of artificially generated random samples from a given distribution (\autoref{sec:AppendixAB}).

Optimal parameters for each model are estimated by maximum likelihood. Then, the best model is selected according to Information Criteria \parencite{anderson2004model}. 
In real languages (this section), models are compared through Akaike Information Criterion (AIC). 
In artificially generated random samples (\autoref{sec:AppendixAB}), the best model is better selected through Bayes Information Criterion (BIC) because the true data generating process is known.
BIC differs from AIC by relying on the assumption that the real distribution is among the tested ones \parencite{wagenmakers2004aic}. 
For a given model, we use the following definitions of these scores \parencite{anderson2004model}
\begin{eqnarray}
AIC & = & - 2\pazocal{L} + 2K \frac{K}{N - K - 1} \nonumber \\
BIC & = & - 2\pazocal{L} + K \log N \label{eq:BIC},
\end{eqnarray}
where $K$ is the number of parameters of the model and $N$ is the sample size. 
\iftoggle{Sonia}{
\textcolor{red}{Sonia: we should clarify if we used the correction for small samples for AIC (as shown in the equation above) or not}
}{}
With respect to AIC, the criterion proposed by Schwarz (BIC) applies a stronger penalty for the number of parameters. \iftoggle{Sonia}{\textcolor{red}{Sonia: still true if we use the corrected AIC?}
}{}

Given that both AIC and BIC are measures of information loss, 
the best model for a sample is the one minimizing the selected score. We aim to find the best model for a sample of $N$ distances $\{d_1,d_2,...,d_i,...,d_N\}$, where $\min(d)$ and $\max(d)$ are, respectively, the minimum and maximum observed distances, and $f(d)$ is the frequency of distance $d$ in the sample. Then the sample size is
\begin{equation*}
N = \sum_{i=1}^{max(d)} f(d_i) = \sum_{d=1}^{max(d)} f(d).
\end{equation*}

The log-likelihood functions of the models are summarized in \autoref{tab:loglikelihoods}. See \autoref{sec:Appendix_log_likelihood} for a derivation of the log-likelihood functions for each model. 

\subsubsection{Parameter estimation }

Maximum likelihood estimation (MLE) algorithms require one to specify the range of variation of the parameters as well as proper initial values. 
It is well-known that MLE methods are highly sensitive to the choice of the starting values, as they may incur local optima when minimizing the minus log-likelihood function \parencite{myung2003tutorial}. Here we explain the criteria used to select the initial value and the range of variation of the parameters, which are summarised in  \autoref{tab:mle_initial} and \autoref{tab:mle_bound} respectively. Let $x_{init}$ be the initial value of parameter $x$. Also, let $\max_i(d)$ be the $i-th$ largest value of $d$ in the sample, so that $\max_1(d)=max(d)$. Similarly,  let $\min_i(d)$ be the $i-th$ smallest value of $d$ in the sample, so that $\min_1(d)=min(d)$. 

\begin{table}[H]
\caption{\label{tab:mle_initial} The initial values of the parameters for maximum likelihood estimation. \iftoggle{Sonia}{}{Here Model 0 refers to model 0.0.}
}
\centering
\begin{tabular}{@{}*{7}{l}}
\toprule
\textbf{Model} & \textbf{$d_{max}$} & \textbf{$q$} & \textbf{$q_1$} & \textbf{$q_2$} & \textbf{$d^*$} & \textbf{$\gamma$} \\
\midrule 
   \textit{0} & $\max(d)$ & - & - & - & - & - \\
   \textit{1} & - & $q_{init}$ & - & - & - & - \\
   \textit{2} & $\max(d)$ & $q_{init}$ & - & - & - & - \\
   \textit{3} & - & - & $q_{1init}$ & $q_{2init}$ & 5 & - \\
   \textit{4} & $\max(d)$ & - & $q_{1init}$ & $q_{2init}$ & 5 & - \\
   \textit{5} & $\max(d)$ & - & - & - & - & $\gamma_{init}$ \\ 
   \textit{6} & - & $q_{init}$ & - & - & 5 & $\gamma_{init}$ \\
   \textit{7} & $\max(d)$ & $q_{init}$ & - & - & 5 & $\gamma_{init}$ \\
\bottomrule
\end{tabular}
\end{table}

The rationale for the choices in  \autoref{tab:mle_initial} and \autoref{tab:mle_bound} is as follows
\begin{itemize}
    \item $d_{max}$. The maximum observed distance is both the starting point and smallest admissible value, while there is no upper bound. 
    \item $q$. In the geometric models (Models 1 and 2), the initial value for $q$, $q_{1init}$, is the maximum likelihood estimator, i.e. the inverse of the mean observed distance $q_{init}=1/mean(d)$. The bounds are set so that $q \in (0,1)$ to avoid values out of the domain of the log-likelihood function. In Models 6 and 7, the initial value of $q$ for the second regime is set to the maximum likelihood estimator $1/mean(d)$ of an ideal geometric distribution, but restricting the mean to distances greater than $d^*$.
    \item $q_1$ and $q_2$. These two parameters are both initialized by first running a linear regression on $\log p(d)$ and $d$, for $d\leq d^*$ in the case of $q_{1init}$, and for $d\geq d^*$ in the case of $q_{2init}$. Then, the respective slopes $\beta_1$ and $\beta_2$ are used to compute the initial values via $q_{1init} = 1 - e^{\beta_1}$ and $q_{2init} = 1 - e^{\beta_2}$. Notice that, as the tail of the distribution is noisy, the estimated slope sometimes results in a 0 or even a positive value for values of $d^*$ very close to $\max(d)$. When that happened, the corresponding $q_{2init}$ was set to its lower bound. As in $q$, the bounds are set so that $q_1, q_2 \in (0,1)$.
    \item $d^*$. The initial value is 5, as suggested by the visual inspection of the plots. The parameter is bounded to vary between $\min_2(d)$ and $\max_2(d)$, based on the minimum requirement on the size of the two regimes (section \ref{subsec:min_size}). Indeed, by setting $d^*$ to either $\min_1(d)$ or to $\max_1(d)$, one of the two regimes would only be composed by one isolated observation, from which no trend can be inferred. Incidentally, the DDm principle, predicts that $\min_2(d)=2$ if $n$ is large enough \parencite{ref:Ferrer-i-Cancho2004}.
    \item $\gamma$. For Model 5, the initial value of the MLE estimator of the exponent of a continuous power-law \parencite{ref:Newman}: 
    \begin{equation*}
        \gamma_{init} = 1 + N \left [ \sum_{i=1}^{N}\frac{d_i}{min(d)} \right ]^{-1}.
    \end{equation*}
    For Models 6 and 7 (where only the first regime follows a zeta distribution), $\gamma_{init}$ is computed over the distances up to $d^*$.
\end{itemize}

\begin{table}[H]
\caption{\label{tab:mle_bound} The lower (low) and upper (up) bounds of the parameters for maximum likelihood estimation. $\epsilon = 10^{-3}$.  
\iftoggle{Sonia}{}{Here Model 0 refers to Model 0.0.}
}
\begin{tabular}{lllllllllllll}
\toprule
    & \multicolumn{2}{c}{$d_{max}$} & \multicolumn{2}{c}{$q$} & \multicolumn{2}{c}{$q_1$} & \multicolumn{2}{c}{$q_2$} & \multicolumn{2}{c}{$d^*$} & \multicolumn{2}{c}{$\gamma$} \\
    \midrule
    \textbf{Model} & \textbf{low} & \textbf{up} & \textbf{low} & \textbf{up} & \textbf{low} & \textbf{up} & \textbf{low} & \textbf{up} & \textbf{low} & \textbf{up} & \textbf{low} & \textbf{up} \\
\midrule
   \textit{0} & $\max(d)$ & $\infty$  & - & - & - & - & - & - & - & - & - & - \\
   \textit{1} & - & - & $\epsilon$ & $1 - \epsilon$ & - & - & - & - & - & - & - & - \\
   \textit{2} & $\max(d)$ & $\infty$  & $\epsilon$ & $1 - \epsilon$ & - & - & - & - & - & - & - & - \\
   \textit{3} & - & - & - & - & $\epsilon$ & $1 - \epsilon$ & $\epsilon$ & $1 - \epsilon$ & $\min_2(d)$ & $\max_2(d)$ & - & - \\
   \textit{4} & $\max(d)$ & $\infty$  & - & - & $\epsilon$ & $1 - \epsilon$ & $\epsilon$ & $1 - \epsilon$ & $\min_2(d)$ & $\max_2(d)$ & - & - \\
   \textit{5} & $\max(d)$ & $\infty$  & - & - & - & - & - & - & - & - & 0 & $\infty$ \\
   \textit{6} & - & - & $\epsilon$ & $1 - \epsilon$ & - & - & - & - & $\min_2(d)$ & $\max_2(d)$ & 0 & $\infty$ \\
   \textit{7} & $\max(d)$ & $\infty$ & $\epsilon$ & $1 - \epsilon$ & - & - & - & - & $\min_2(d)$ & $\max_2(d)$ & 0 & $\infty$ \\
\bottomrule
\end{tabular} 
\end{table}

\subsubsection{Maximum likelihood estimation (MLE) }

We considered two  
MLE methods
in \texttt{R}: \texttt{mle()} from \texttt{stats2} and \texttt{mle2()} from the \texttt{bbmle} package \parencite{Bolker2007a}. The base \texttt{R} implementation, \texttt{mle()}, may explore values out of the specified bounds thus resulting in errors. Where this is the case, we resort to the enhanced, more robust version of the optimizer,  \texttt{mle2()}, which is able to return a result even if the algorithm does not reach convergence.
Both \texttt{mle2a()} and \texttt{mle2()} optimize on a continuous space. Hence, for the discrete parameters, i.e. $d^*$ and $d_{max}$, we retrieved their most likely value by exhaustively exploring all values included between their theoretical bounds. In this way, we also decrease the complexity of MLE by reducing the number of parameters to be optimized through the call to
\texttt{mle()} or
\texttt{mle2()}. Thus, for each value of $d^*$ (and $d_{max}$ in the right-truncated models) we optimized the remaining parameters, and finally selected the parameters combination resulting in the highest log-likelihood. 

\subsubsection{Requirements for two-regime models}\label{subsec:min_size}

\begin{figure}[!htbp]
\centering
\includegraphics[width = 0.5\textwidth]{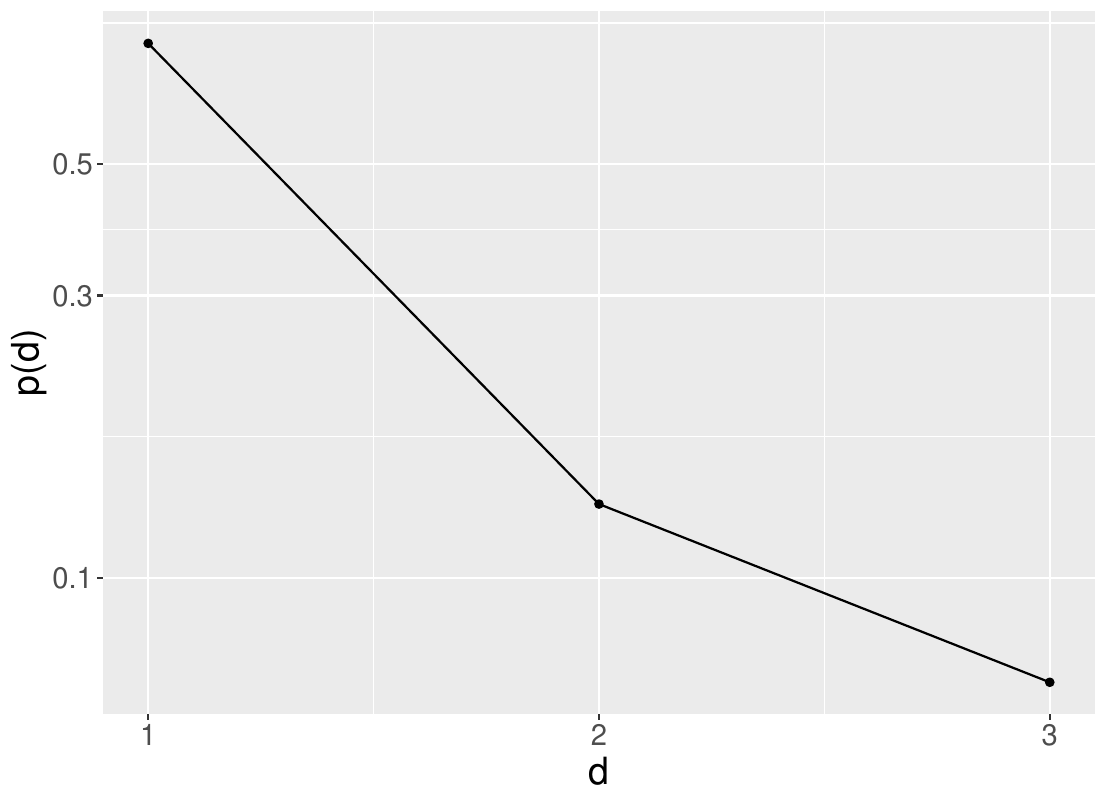}
\caption{Syntactic dependency distance distribution of sentences with 4 words in Italian, annotated according to SUD. Only three unique values of $d$ have been observed. } \label{fig:min_size}
\end{figure}

In order to fit a double-regime model to a data sample, we need 
$N \geq 3$.
In fact, at least two points are needed in order to infer a speed of probability decay within a regime, meaning that each regime has to contain at least 2 distinct observations. Given that the value assigned to the break-point is common to the two regimes, this results in a requirement of $N \geq 3$.
See \autoref{fig:min_size} for an example of this scenario, displaying the distribution of syntactic dependency distances for sentences of 4 words in Italian, annotated according to SUD. Notice that this requirement directly implies that sentences with $n<4$ are excluded from the model selection procedure.

\subsubsection{Representativeness}
\label{subsec:representativity}

When performing model selection on sentences of specific lengths from a certain language, we obtain a set of best models, one for each sentence length. To summarize this information and obtain a single best model for each language, we consider the most frequent best model within that language. However, this raises the concern of whether all best models are equally reliable, as some of them are estimated on a single sentence. For instance, very long sentences, which are normally rare, are thus likely to be underrepresented in the data. On the other hand, setting a single specific threshold on the minimum number of sentences required for a sentence length to be included in the voting procedure would mistakenly hide important aspects of the analysis.
In fact, the 
suitable threshold should depend on
sentence length itself. Consider a very long sentence, composed of 50 words, and a very short one, of only 4 words.
While -- keeping fixed the syntactic structure -- the first one could appear with $50!$ different re-orderings, the second one could only be written in $4!$ possible ways. Thus, a single sentence observed for $n=4$ is much more representative (as the expected variability in dependency distance is lower) for the whole length category than a single one observed for $n=50$. For this reason, we report the most frequent best models both when no threshold is set (\autoref{tab:best_PUD_PSUD_fixed}) and for increasing representativeness threshold (\autoref{fig:threshold}).

\subsection{The \texorpdfstring{$\Omega$}{} optimality score}\label{subsec:omega}
$\Omega$ is a recently introduced optimality score for the closeness of syntactic dependency distances, which integrated normalization with respect to both a minimum and a random baseline \parencite{Omega}. The score is defined as
\begin{equation*}
    \Omega = \frac{D_{rla}-D}{D_{rla}-D_{min}},
\end{equation*}
where $D$ is the observed sum of dependency distances in a sentence, $D_{rla}$ is the expected sum of dependency distances in a uniformly random linear arrangement of its words \parencite{ref:Ferrer-i-Cancho2004,Ferrer2019a}, i.e.
\begin{equation}
D_{rla} = \frac{n^2 - 1}{3}
\end{equation}
and $D_{min}$ is the sum of dependency distances in a minimum linear arrangement of the words \parencite{Shiloach,Esteban2017a}. Both baselines assume that the network structure is fixed. $D$ and $D_{min}$ are computed using the \texttt{python} interface of the Linear Arrangement Library  \parencite{Alemany2021b}.

Positive values of $\Omega$ indicate that syntactic dependency distances in the sentence are shorter than one would expect from picking uniformly at random among all the possible $n!$ orderings. The maximum, $\Omega=1$, is reached when $D = D_{min}$. Conversely, a negative value indicates that distances are being maximized, as they are higher than expected in a random shuffling of words in a sentence. When word order is random, $\Omega$ will take values tending to 0.
$\left< \Omega \right>$ is the average value of $\Omega$ over individual sentences.

\section{Results}
\label{sec:Results}
We fit the models introduced in the Section \ref{sec:Models} to a parallel collection of texts from 20 languages called PUD, that has been annotated with syntactic dependencies as in \autoref{fig:example_tree}. To control for annotation style we consider two variants, PUD with the original annotation style \parencite{nivre-etal-2017-universal} and PSUD, that follows the alternative SUD annotation style \parencite{ref:SUD}. Refer to section \ref{sec:Materials} for further details on the data, and to section \ref{subsec:model_selection} for a complete description of the model selection procedure.

This section is organized as follows. First, we report on the best models (Section \ref{subsec:results_model_selection}), the break-points of the two-regime models (Section \ref{subsec:break_point}) and the relationship between slope parameters ($q_1$ and $q_2$) for each language (Section \ref{subsec:speed_of_decay}), both by considering fixed and mixed sentence lengths. We define representativeness threshold, shortly representativeness, as the minimum number of distinct sentences with a certain length for such length to be included in model selection (a further justification of this threshold is found in Section \ref{subsec:representativity}. Section \ref{subsec:results_model_selection} 
\iftoggle{Sonia}{
\textcolor{red}{only this section???}} {}
investigates the robustness of conclusions with respect to sample representativeness. Detailed tables of the estimated parameters, Akaike Information Criterion (AIC) scores, and AIC differences for both collections can be found in \autoref{sec:AppendixB}. Second, we will investigate the relationship between the best model and the degree of optimality of syntactic dependency distances on sentences of fixed length (Section \ref{subsec:best_model_versus_optimality}. Notice that we often refer jointly to Models 3 and 4 (6 and 7) as 3-4 (6-7), given that they model the same probability distribution with or without a right-truncation point. 


\subsection{Model selection}
\label{subsec:results_model_selection}

\begin{figure}[!htbp]
\centering
    \includegraphics[width = 0.85\textwidth]{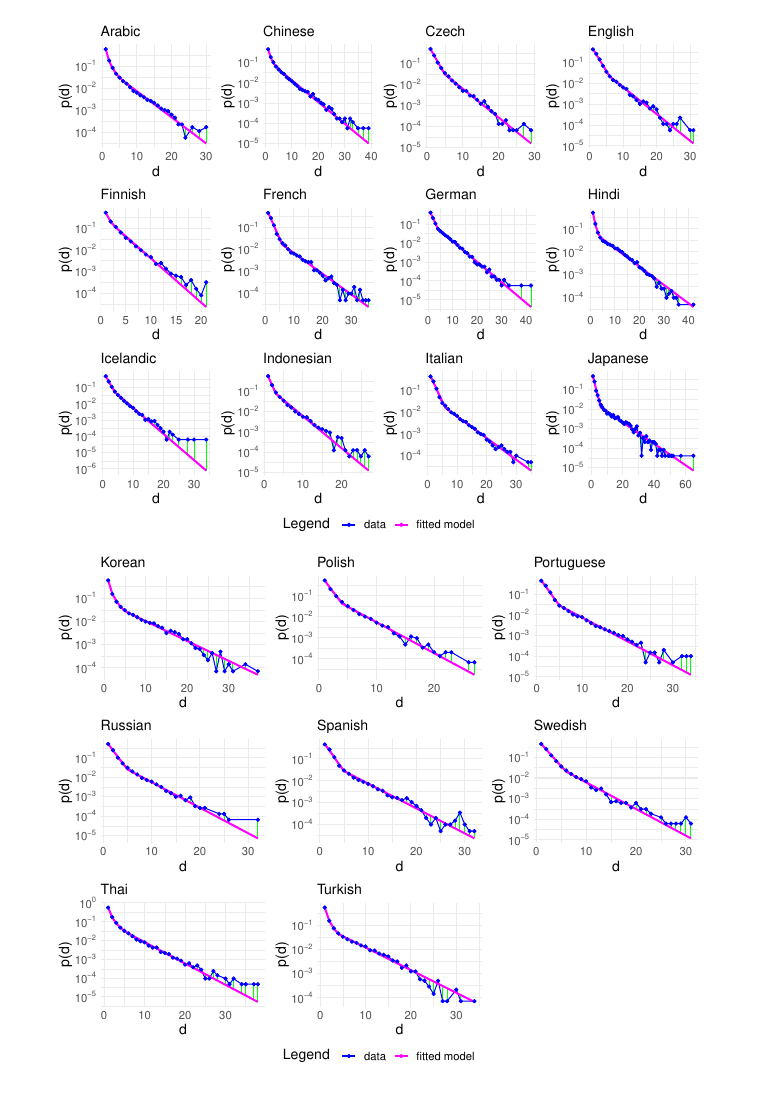}
    \caption{$p(d)$, the probability that a dependency link is formed between words at distance $d$ according to the data and the best model for every language in PUD.} \label{fig:PUD_best_fitted}
\end{figure}

The best model to describe syntactic dependency distances independent of sentence length is composed of two regimes in every language and collection (\autoref{tab:best_PUD_PSUD}). 
Models 3-4 dominate over Models 6-7, with 13/20 languages in PUD and 11/20 in PSUD having Model 3 or 4 as the best one.
We find overall agreement between the two annotation styles, both in terms of best model and in terms of right-truncation. The exceptions to this agreement are Indonesian and Japanese -- for which PUD yields an exponential decay in the first regime, while PSUD identifies a power-law one -- and Chinese, English, and Italian, where the best model in PUD and PSUD differs by right truncation. In \autoref{fig:PUD_best_fitted}, we show how the best models in PUD are able to accurately capture the bulk of the distribution, with some variability left in the tail. 
The equivalent figure for PSUD can be found in \autoref{sec:AppendixB}.

\begin{table}[H]
\caption{Best model for the distribution of syntactic dependency distances in sentences of mixed lengths for every language and collection. Models 3-4 are marked with pink and Models 6-7 with blue to ease visualization.}
\centering

        \begin{tabular}{@{}*{3}{l}}
        \toprule
        \textbf{Language} & \textbf{PUD} & \textbf{PSUD}\\
        \midrule        

Arabic &   \cellcolor{blue!25}7 &   \cellcolor{blue!25}7 \\ 
  Chinese &  \cellcolor{blue!25}6 & \cellcolor{blue!25}7 \\ 
  Czech &   \cellcolor{pink}3 &   \cellcolor{pink}3 \\ 
  English &   \cellcolor{pink}3 & \cellcolor{pink}4 \\
  Finnish &   \cellcolor{blue!25}6 &   \cellcolor{blue!25}6 \\ 
  French &   \cellcolor{pink}4 &   \cellcolor{pink}4 \\ 
  German &   \cellcolor{pink}3 &   \cellcolor{pink}3 \\ 
  Hindi &   \cellcolor{blue!25}7 &   \cellcolor{blue!25}7 \\ 
  Icelandic &   \cellcolor{pink}3 &   \cellcolor{pink}3 \\ 
  Indonesian &   \cellcolor{pink}3 & \cellcolor{blue!25}7 \\

        \bottomrule    
        \end{tabular}
\hspace{0.025\textwidth}
        \begin{tabular}{@{}*{3}{l}}
        \toprule
        \textbf{Language} & \textbf{PUD} & \textbf{PSUD}\\
        \midrule

Italian & \cellcolor{pink}4 &\cellcolor{pink}3 \\
Japanese & \cellcolor{pink}4 & \cellcolor{blue!25}7 \\ 
Korean & \cellcolor{blue!25}7 & \cellcolor{blue!25}7\\ 
Polish &  \cellcolor{pink}3 &   \cellcolor{pink}3 \\
Portuguese &  \cellcolor{pink}3 &   \cellcolor{pink}3\\
Russian & \cellcolor{pink}3 &   \cellcolor{pink}3 \\ 
Spanish & \cellcolor{pink}4 &   \cellcolor{pink}4 \\ 
Swedish & \cellcolor{pink}3 &   \cellcolor{pink}3 \\ 
Thai & \cellcolor{blue!25}6 &   \cellcolor{blue!25}6 \\
Turkish & \cellcolor{blue!25}7 & \cellcolor{blue!25}7\\

        \bottomrule 
        \end{tabular}

    \label{tab:best_PUD_PSUD} 
\end{table}

\begin{table}[H]
\caption{Most voted best model for the distribution of syntactic dependency distances in sentences of fixed lengths, for every language and collection. The most voted best model is computed aggregating models by type, thus counting together the occurrences in which Models 3-4 (Models 6-7) are the best. Models 3-4 are marked with pink, Model 5 with yellow, and Models 6-7 with blue to ease visualization.}

\centering
        \begin{tabular}{@{}*{3}{l}}
        \toprule
        \textbf{Language} & \textbf{PUD} & \textbf{PSUD} \\
        \midrule
Arabic & \cellcolor{yellow}5 & \cellcolor{yellow}5 \\ 
  Chinese & \cellcolor{yellow}5 & \cellcolor{yellow}5 \\ 
  Czech & \cellcolor{pink}3-4 & \cellcolor{pink}3-4 \\ 
  English & \cellcolor{pink}3-4 & \cellcolor{pink}3-4 \\ 
  Finnish & \cellcolor{blue!25}6-7 & \cellcolor{blue!25}6-7 \\ 
  French & \cellcolor{pink}3-4 & \cellcolor{pink}3-4 \\ 
  German & \cellcolor{pink}3-4 & \cellcolor{pink}3-4 \\ 
  Hindi & \cellcolor{blue!25}6-7 & \cellcolor{blue!25}6-7 \\ 
  Icelandic & \cellcolor{pink}3-4 & \cellcolor{pink}3-4 \\ 
  Indonesian & \cellcolor{pink}3-4 & \cellcolor{yellow}5 \\ 

        \bottomrule
        \end{tabular}
\hspace{0.025\textwidth}
        \begin{tabular}{@{}*{3}{l}}
        \toprule
        \textbf{Language} & \textbf{PUD} & \textbf{PSUD} \\
        \midrule
Italian & \cellcolor{pink}3-4 & \cellcolor{pink}3-4 \\ 
  Japanese & \cellcolor{pink}3-4 & \cellcolor{pink}6-7 \\ 
  Korean & \cellcolor{blue!25}6-7 & \cellcolor{blue!25}6-7 \\ 
  Polish & \cellcolor{pink}3-4 & \cellcolor{blue!25}5 \\ 
  Portuguese & \cellcolor{pink}3-4 & \cellcolor{pink}3-4 \\ 
  Russian & \cellcolor{pink}3-4 & \cellcolor{yellow}5 \\ 
  Spanish & \cellcolor{pink}3-4 & \cellcolor{pink}3-4 \\ 
  Swedish & \cellcolor{pink}3-4 & \cellcolor{pink}3-4 \\ 
  Thai & \cellcolor{yellow}5 & \cellcolor{yellow}5 \\ 
  Turkish & \cellcolor{blue!25}6-7 & \cellcolor{blue!25}6-7 \\

        \bottomrule
        \end{tabular}

    \label{tab:best_PUD_PSUD_fixed} 

\end{table}

\begin{figure}[!htbp]
\centering
    \includegraphics[width = 0.85\textwidth]{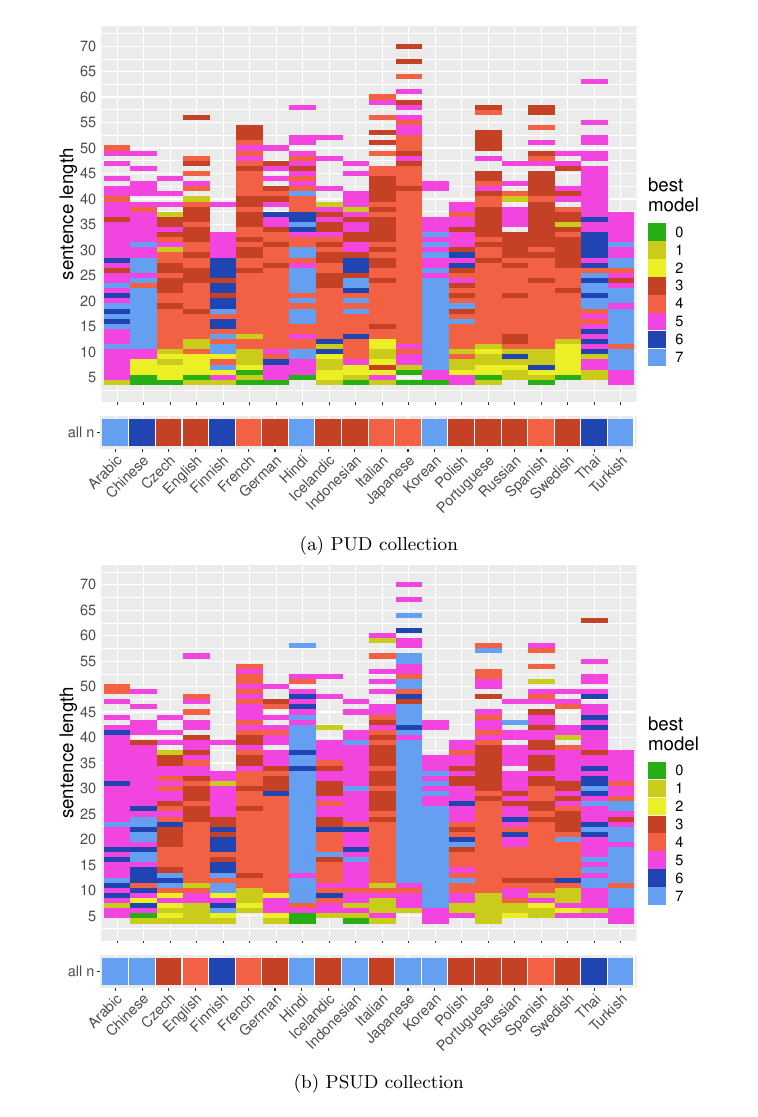}
    \caption{\label{fig:best_model} Distribution of best model for each sentence length on top, with reference to the best model on mixed sentence lengths at the bottom. (a) PUD collection. (b) PSUD collection. In both (a) and (b) the empty tiles mark lengths for which no sentence was observed, or on which model selection was not performed given the minimum requirement to fit a double-regime model, described in section \ref{subsec:min_size}. 
    \iftoggle{Sonia}{\textcolor{red}{Model 0 here could be Model 0 or Model 0.1}}{Here Model 0 refers to Model 0.0.} } 
\end{figure}

\begin{figure}[!htbp]
  \centering
    \includegraphics[width = 0.85\textwidth]{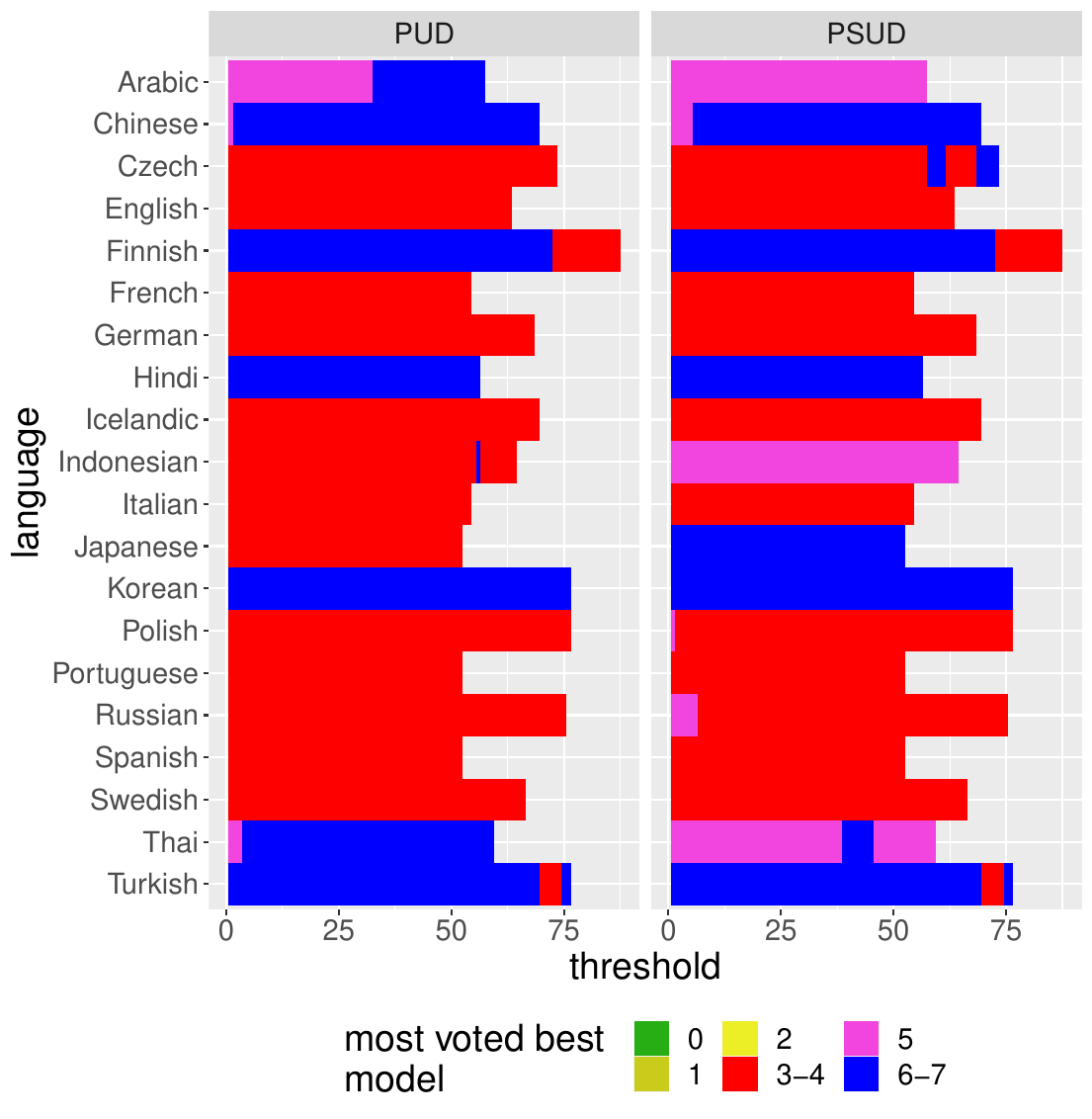}
    \caption{\label{fig:threshold} Most voted best model type across sentence lengths for increasing representativeness threshold. When no threshold is set (1 minimum sentence), we get the scenario displayed in \autoref{fig:best_model}. Ties are counted in favour of models without two regimes. \iftoggle{Sonia}{\textcolor{red}{Model 0 here could be Model 0 or Model 0.1}}{Here Model 0 refers to Model 0.0.} }
\end{figure}

The best model for sentences of fixed lengths shows some variability for short and long sentences (\autoref{fig:best_model}). Nevertheless, a double regime model is the most frequent best one across sentence lengths in 17/20 languages in PUD (including a tie between Model 5 and Models 6-7 in Chinese), and in 14/20 languages in PSUD (\autoref{tab:best_PUD_PSUD_fixed}). 
Within the languages for which a two-regime model is the best one, 
Models 3-4 win in 13/17 languages in PUD, and in 9/14 in PSUD. Once again, we find high consistency between annotation styles, with the exceptions of Indonesian, Polish, and Russian, for which PSUD yields Model 5 as the most frequent best one (while PUD yields models 3-4), and Japanese, for which PUD and PSUD differ in the type of two-regime model. Finally, Model 5 is the most frequent best one in both collections for Arabic, Chinese, and Thai. However, \autoref{fig:threshold} shows how the most voted best model ceases to be Model 5 in some instances of both PUD and PSUD when the representativeness of a sentence length is taken into account. 
The only languages in which Model 5 is consistently the most frequent best one even after imposing an arbitrary high threshold are Thai, Indonesian, and Arabic in PSUD. Arabic shows a border-line behaviour in PUD, with Model 5 being consistently the most voted only up to a certain threshold value.
Finally, a comparison of the actual distribution against the best model in sentences of fixed characteristic length is shown in \autoref{sec:AppendixC}.

\subsection{The break-point}
\label{subsec:break_point}

\begin{figure}[!htbp]
  \centering
     \includegraphics[width = 0.85\textwidth]{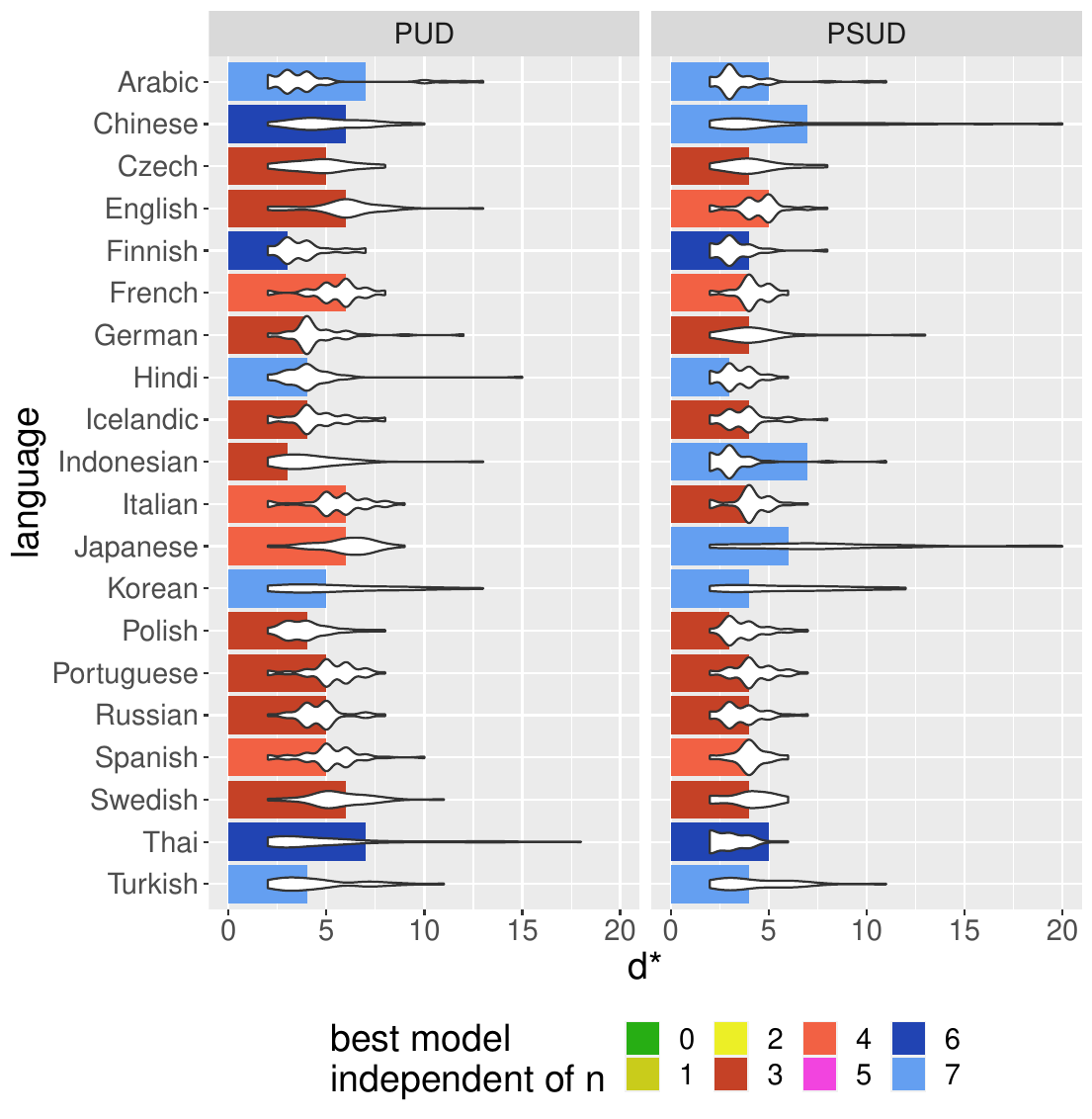}
     \caption{\label{fig:dstar_violinplot} 
    Value of $d^*$ for mixed sentence lengths (bars) in each language and collection, and its distribution across fixed sentence lengths (violin plots), color-coded by best model independent of sentence length (namely the best model estimated on sentences of mixed lengths).
        \iftoggle{Sonia}{\textcolor{red}{Model 0 here could be Model 0 or Model 0.1}}{Model 0 refers to Model 0.1 in the context of mixed sentence lengths.} 
    }
\end{figure}

When looking at languages globally, meaning considering jointly sentences of any length, we find that the break-point $d^*$ always takes small values -- ranging between 2 and 7 -- and has a quite small standard deviation (\autoref{fig:dstar_violinplot} and \autoref{tab:dstar_summary}), meaning that its value is similar across languages. This is especially true for Models 3-4 and the PSUD collection: out of 11 languages having Models 3-4 as the best ones in this collection, 9 have an estimated break-point at $d^*=4$ (\autoref{fig:dstar_violinplot}). In PUD these models have an average $d^*$ value of 5, but with some more variability across languages. In both types of two regime models median and mean values are virtually the same, independently of annotation style, providing additional evidence for the low variance of $d^*$ (\autoref{tab:dstar_summary}). Checking the distribution of $d^*$ within a language allows us to verify whether global values (found when mixing sentence lengths), namely the bars in  \autoref{fig:dstar_violinplot}, are good approximations of the break-points actually observed in real sentences of any fixed length. We display the distribution of $d^*$ across sentence lengths for each language in the same figure as a violin plot. Once again the median is very close to the mean in almost every combination of two-regime model and annotation style -- with the exception of Models 6-7 in PSUD -- further supporting $H_2$ (\autoref{tab:dstar_summary_fixed_n}).
Then, notice that where Models 3-4 are the best we observe relatively narrow distributions, skewed towards low values and showing one or a few modes (\autoref{fig:dstar_violinplot}). In particular, the global value of $d^*$ is virtually always found in correspondence of one of these modal values, confirming its representativeness for the whole language. 
Considering that sentences can reach up to a minimum of 37 (Turkish) and a maximum of 70 words (Japanese) (\autoref{tab:PUD_summary}) the observed variation ranges in Models 3-4 are quite small, with values going up to roughly $d^* = 13$. On the other hand, within languages for which Models 6-7 are the best when mixing sentence lengths, the distribution of $d^*$ across different sentence lengths is generally flatter, especially in PSUD. Even where values are centered around a mode, this does not correspond with the break-point estimated globally, with the exception of Hindi. Thus, it appears like the global break-points estimated in Models 3-4 are good approximations of the values observed within the language, while estimates of $d^*$ in Models 6-7 are less reliable as representations of the actual break-point if there is any. 

\begin{table}[!htbp]
    \caption{Summary statistics of the $d^*$ parameter, by annotation style and type of two-regime model, estimated from model selection on sentences of mixed lengths. The summary is computed over languages where Models 3-4 are the best, where Models 6-7 are the best, and over all languages where a double-regime model is the best (Models 3-4-6-7). Thus, sample size is measured in number of languages. $s$ stands for sample size, $sd$ stands for standard deviation. \label{tab:dstar_summary} }
\centering    
     \begin{tabular}{@{}*{10}{l}}
     \toprule                              
     & \textbf{Models} & \textbf{$s$} & \textbf{Min.} & \textbf{1st Qu.} & \textbf{Median} & \textbf{Mean} & \textbf{3rd Qu.} & \textbf{Max.} & \textbf{$sd$} \\
     \midrule
     \multirow{3}{*}{PUD} 
& 3-4 & 13 & 3.00 & 4.00 & 5.00 & 5.00 & 6.00 & 6.00 & 1.00 \\ 
 & 6-7 & 7 & 3.00 & 4.00 & 5.00 & 5.14 & 6.50 & 7.00 & 1.57 \\ 
 & 3-4-6-7 & 20 & 3.00 & 4.00 & 5.00 & 5.05 & 6.00 & 7.00 & 1.19 \\ 

     \midrule
     \multirow{3}{*}{PSUD} 
& 3-4 & 11 & 3.00 & 4.00 & 4.00 & 4.00 & 4.00 & 5.00 & 0.45 \\ 
 & 6-7 & 9 & 3.00 & 4.00 & 5.00 & 5.00 & 6.00 & 7.00 & 1.41 \\ 
 & 3-4-6-7 & 20 & 3.00 & 4.00 & 4.00 & 4.45 & 5.00 & 7.00 & 1.10 \\

     \bottomrule
     \end{tabular}
\end{table}


\begin{table}[!htbp]
\caption{Summary statistics of $d^*$ parameter, by collection and type of two-regime model, estimated from model selection on sentences of fixed lengths. The summary is computed over sentence lengths and languages where Models 3-4 are the best, where Models 6-7 are the best, and over all languages and sentence lengths where a double-regime model is the best (Models 3-4-6-7). Thus, sample size is measured in number of distinct sentence lengths. $s$ stands for sample size, $sd$ stands for standard deviation. }
\centering
    \begin{tabular}{@{}*{10}{l}}
     \toprule
     & \textbf{Models} & $s$ & \textbf{Min.} & \textbf{1st Qu.} & \textbf{Median} & \textbf{Mean} & \textbf{3rd Qu.} & \textbf{Max.} & \textbf{$sd$} \\
     \midrule
    \multirow{3}{*}{PUD}  
& 3-4 & 431 & 2.00 & 4.00 & 5.00 & 5.37 & 6.00 & 13.00 & 1.43 \\ 
 & 6-7 & 134 & 2.00 & 4.00 & 6.00 & 6.28 & 7.00 & 18.00 & 3.03 \\ 
 & 3-4-6-7 & 565 & 2.00 & 4.00 & 5.00 & 5.59 & 6.00 & 18.00 & 1.97 \\ 
    \midrule
    \multirow{3}{*}{PSUD} 
&  3-4 & 297 & 3.00 & 4.00 & 4.00 & 4.32 & 5.00 & 13.00 & 1.11 \\ 
 &  6-7 & 190 & 2.00 & 3.00 & 5.00 & 6.21 & 8.00 & 20.00 & 3.73 \\ 
 &  3-4-6-7 & 487 & 2.00 & 4.00 & 4.00 & 5.06 & 5.00 & 20.00 & 2.65 \\ 
    \bottomrule
    \end{tabular}
    \label{tab:dstar_summary_fixed_n} 
\end{table}

\subsection{Speed of decay}
\label{subsec:speed_of_decay}

Recall that $q_1$ and $q_2$ are the slope parameters of Models 3-4, which quantify the speed of probability decay. For each language in which a two-regime model is the best, we consider $q_1$, $q_2$, and their ratio $q_1/q_2$, where the latter quantity is computed to establish which slope is steeper. It has been suggested that the probability decay is slower in the 2nd regime \parencite{ref:Ferrer-i-Cancho2004,Ferrer2017d}. When Models 6-7 are the best models, we estimate $q_1$ of the first regime by fitting the corresponding double exponential model (Model 3 or 4).
When we refer to a slope, we refer to its absolute value.

Where the best model has two regimes, the estimated slope parameters for each regime are fairly similar across languages (\autoref{fig:mixed_n_slopes} and \autoref{tab:mixed_n_slopes}). In addition, notice that the ratio $q_1/q_2$ is larger than 1 for every language and annotation style, and that $q_1$ and $q_2$ have a quite small standard deviation {(\autoref{tab:mixed_n_slopes}). Standard deviation values are practically the same for the two parameters, but $q_2$ takes much lower values, meaning that it is relatively more variable than $q_1$. Moreover -- as in the case of the break-point parameter -- median and mean values are virtually the same, for both $q_1$ and $q_2$ and independently of annotation style. The slope estimated in the first regime in PUD is significantly lower than the one estimated in PSUD (\autoref{fig:mixed_n_slopes} (a)). Moreover, the estimated slopes show a clear pattern, with probability in the first regime consistently decaying faster compared to the second one. This pattern holds for the overwhelming majority of sentence lengths within a language, with a few exceptions found for very short sentences (\autoref{fig:fixed_n_slopes}).

\begin{figure}[!htbp]
\centering
    \includegraphics[width = 0.9\textwidth]{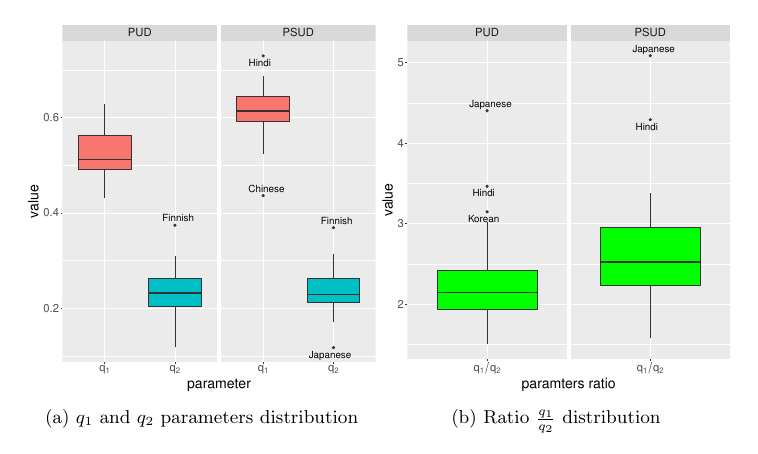}
    \caption{Distribution of slope parameters $q_1$ and $q_2$ and their ratio. Isolated points are labelled with the corresponding language.} \label{fig:mixed_n_slopes}
\end{figure}


\begin{table}[!htbp]
\caption{Summary statistics of $q_1$ and $q_2$ parameters and their ratio ($q_1/q_2$) for model selection on sentences of mixed lengths, by annotation style (referred to as collection). Statistics are computed over all sentence lengths and languages for which a double-regime model is the best. $sd$ stands for standard deviation.
\label{tab:mixed_n_slopes}}
\centering
   \begin{tabular}{@{}*{9}{l}}
   \toprule
    & \textbf{Collection} & \textbf{Min.} & \textbf{1st Qu.} & \textbf{Median} & \textbf{Mean} & \textbf{3rd Qu.} & \textbf{Max.} & \textbf{$sd$} \\
    \midrule
    \multirow{2}{*}{$q_1$}  
&  PUD & 0.43 & 0.49 & 0.51 & 0.52 & 0.56 & 0.63 & 0.05 \\ 
&  PSUD & 0.44 & 0.59 & 0.61 & 0.61 & 0.65 & 0.73 & 0.06 \\

    \midrule
    \multirow{2}{*}{$q_2$}  
& PUD & 0.12 & 0.20 & 0.23 & 0.24 & 0.26 & 0.37 & 0.06 \\ 
 &  PSUD & 0.12 & 0.21 & 0.23 & 0.24 & 0.26 & 0.37 & 0.05 \\

    \midrule    
    \multirow{2}{*}{$q_1/q_2$}  
& PUD & 1.50 & 1.94 & 2.15 & 2.32 & 2.42 & 4.40 & 0.70 \\ 
 &  PSUD & 1.58 & 2.24 & 2.52 & 2.75 & 2.95 & 5.09 & 0.79 \\

    \bottomrule    
    \end{tabular} 

\end{table}

\begin{figure}[!htbp]
  \centering
     \includegraphics[width = 0.75\textwidth]{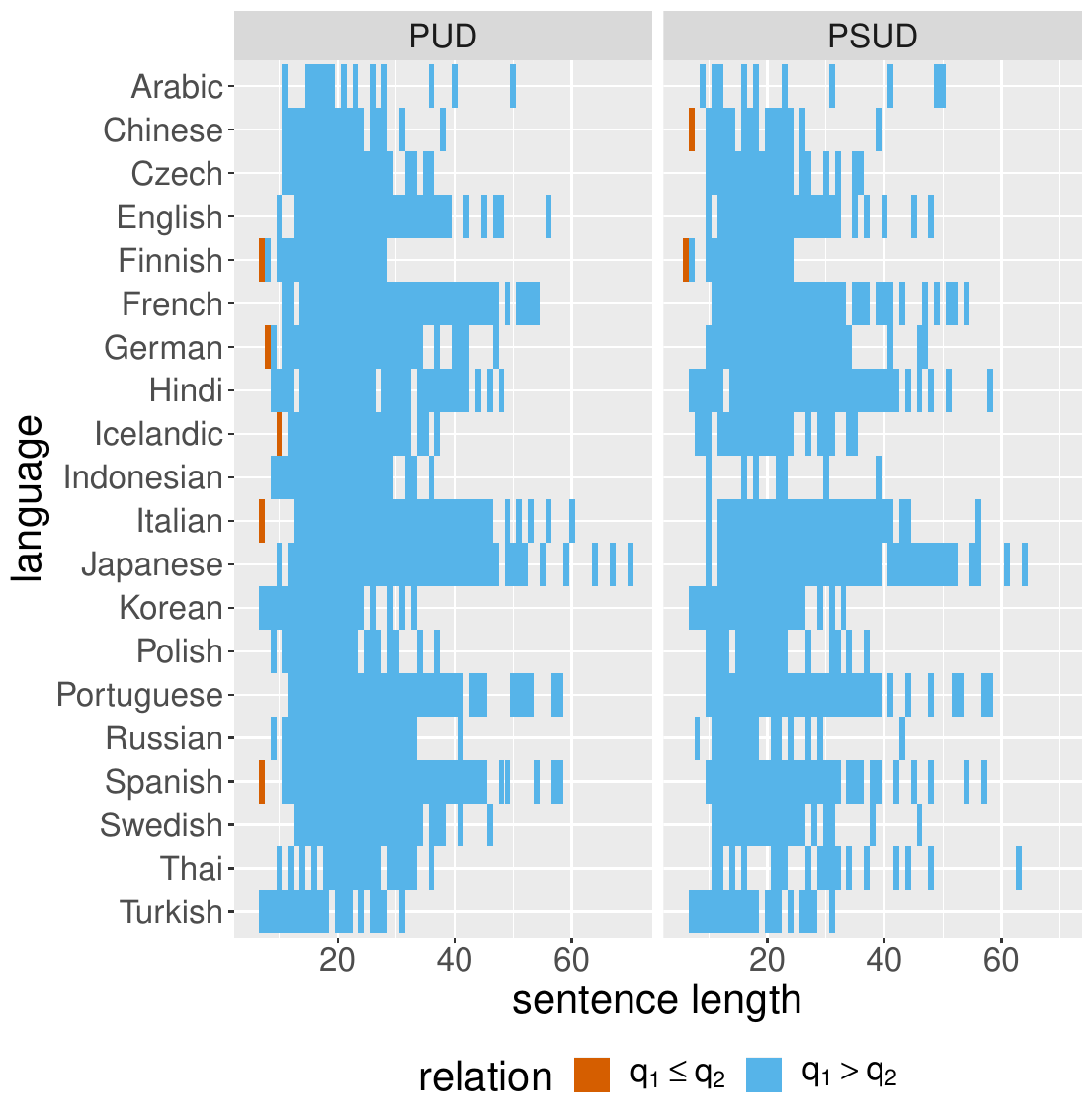}
     \caption{\label{fig:fixed_n_slopes}
    Relation between slope parameters $q_1$ and $q_2$ estimated from model selection on fixed sentence lengths. Lengths for which $q_1 \leq q_2$ are colored in red, while those for which $q_1 > q_2$ are colored in blue. Where the best model was 6 (7), the first slope was approximated by fitting Model 3 (4) with the original value of $d^*$. The empty tiles indicate lengths for which no sentence was observed, a two-regime model was not the best one, or on which model selection was not performed given the minimum requirement on the number of observed distance values to fit a double-regime model, described in section \ref{subsec:min_size}.  
    }
\end{figure}

\subsection{The best model versus the optimality of syntactic dependency distances}
\label{subsec:best_model_versus_optimality}

$\Omega$ is a new closeness score for syntactic dependency distances. The higher its value, the closer the syntactically related words. Refer to section \ref{subsec:omega} for further details on its properties and computation. The score takes positive values when syntactic dependency distances are minimized, negative values when they go against minimization, and values around 0 when there is no pressure in either direction \parencite{Omega}. Let $\left<\Omega \right>$ be the average value of $\Omega$ over all sentences with a given length in a language. 
See \autoref{fig:PUD_omega} and \autoref{fig:PSUD_omega} for the best model for each sentence length (a) and the corresponding value of $\left <\Omega \right>$ (b), for PUD and PSUD respectively. First, in sentences of a very few words, the best model is either Model 0 or one with a single regime, and the values of the optimality score signal the coexistence of the three possible systems: anti-DDm (orange tiles), no bias (white tiles), and pro-DDm (purple tiles). Given the definition of the score, we expect that, under the assumption that Model 0 is the real distribution, $\left <\Omega \right>$ will take values around 0, as both situations underlie random word ordering. This expectation is met in 6/8 instances, as displayed in \autoref{tab:omega_0}, and as suggested by the correspondence between white tiles in (b) and green tiles in (a). The two exceptions are Korean in PSUD and Polish in PUD, for which the best model is Model 5. Then, for sentences longer than 5-6 words, $\left <\Omega \right>$ indicates that distances in syntactic structures are always minimized, which is mirrored in the disappearance of Model 0 and the predominance of the single regime models. Finally, as pressure for minimization further increases with sentence length, these simpler models are progressively replaced by the models with two regimes.


\begin{table}[!htbp]
\caption{Estimated best model on fixed sentence in collections, languages, and sentence lengths for which $|\left<\Omega \right> - \epsilon| \leq 0$, with $\epsilon = 0.1$. $\left<\Omega \right>$ is the average value of $\Omega$ over all sentences with a given length in a language. }
\centering
   \begin{tabular}{lllrl}
\toprule
    \textbf{Collection} & \textbf{Language}  & $n$ & $\left<\Omega\right>$ & \textbf{Best model} \\
\midrule
  PUD & Korean &   4 & $-0.05$ &   0 \\ 
   & Czech &   4 & 0.00 &   0 \\ 
   & French &   4 & 0.00 &   0 \\ 
   & Spanish &   4 & 0.00 &   0 \\ 
   & Polish &   4 & 0.08 &   5 \\ 
   & Chinese &   4 & 0.08 &   0 \\ 
  PSUD & Korean &   4 & $-0.10$ &   5 \\ 
   & Hindi &   4 & 0.00 &   0 \\
\bottomrule    
    \end{tabular}
    \label{tab:omega_0} 
\end{table}

\begin{figure}[!htbp]
\centering
    \includegraphics[width = 0.85\textwidth]{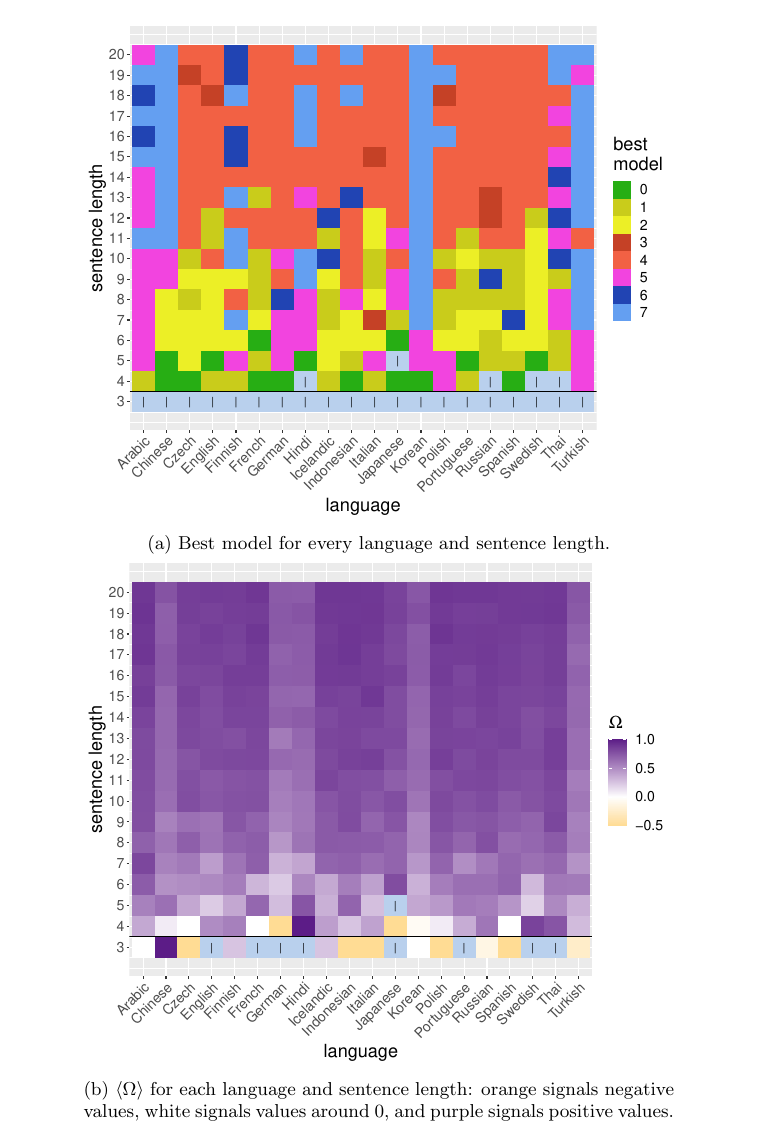}
    \caption{Relation between $\Omega$ score and best model in PUD. The barred gray cells indicate the sentence lengths which have not been observed, or that were excluded from model selection according to the representativeness threshold.  Sentence lengths are cut at $n=20$ to ease visualization. 
    \iftoggle{Sonia}{\textcolor{red}{Model 0 here could be Model 0 or Model 0.1}}{Model 0 refers to Model 0.0.} 
    In (b), orange signals negative
values, white signals values around 0, and purple signals positive values.
   \iftoggle{Sonia}{
   \textcolor{red}{Sonia: within the PDF of the figure remove ": orange signals negative
values, white signals values around 0, and purple signals positive values; that text should be moved to the actual legend of the figure; this figure should be transformed into a proper figure with two subfigures}}
{}
} \label{fig:PUD_omega}
\end{figure}

\begin{figure}[!htbp]
\centering
    \includegraphics[width = 0.85\textwidth]{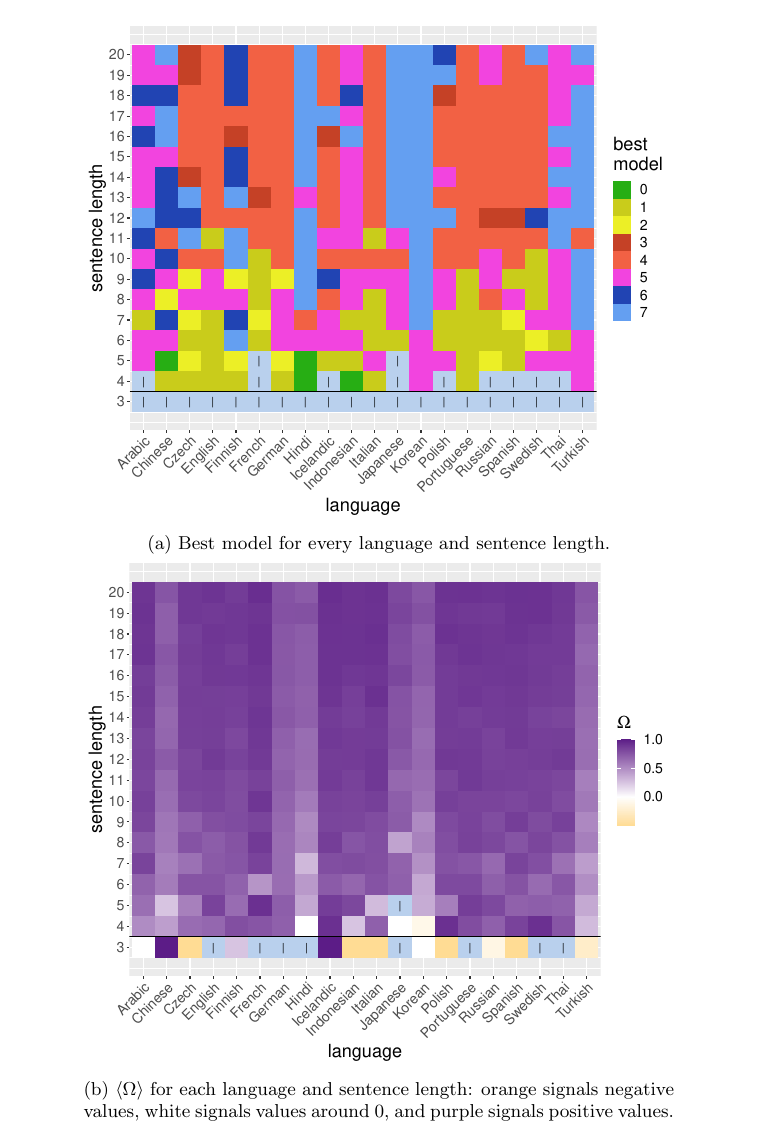}
    \caption{\label{fig:PSUD_omega} Relation between $\Omega$ score and best model in PSUD. The format is the same as in \autoref{fig:PUD_omega}.}
\end{figure}

\section{Discussion}
\label{sec:Discussion}

First, we focus on the two hypotheses object of study, namely that syntactic dependency distances are distributed following two exponential regimes ($H_1$), and that the break-point shows low variation across languages ($H_2$). Our results provide strong evidence for both hypotheses in a large group of languages, mainly Indo-European, consistently across annotation styles. Second, we 
reflect on the parameters yielding the best fit and pay attention to the greater steepness of the first regime with respect to the second one, and the homogeneity of the estimated slopes across languages.
Finally, we discuss the relation between the best estimated model and the closeness of syntactic dependencies as captured by the optimality score $\Omega$ \parencite{Omega}, and summarize the effect of annotation style.

\subsection{The reality of two regimes}\label{subsec:two_regimes}

\subsubsection{The shape of the distribution}\label{subsec:distribution}
As it is often the case, the path to the truth seems to lie in the middle. We could neither generalize to all languages hypothesis $H_1$ (supported by 13/20 languages in PUD and by 11/20 in PSUD), first advanced by \textcite{ref:Ferrer-i-Cancho2004}, nor fully corroborate the finding that dependency distances are power-law distributed as reported for Chinese \parencite{ref:HaitaoLiu}. However, we provided evidence for a possible explanation integrating both: a two-regime model in which the first regime is either exponential or power-law distributed, and the second one follows an exponential decay. A two-regime model is found in all languages when mixing sentences of different lengths (\autoref{tab:best_PUD_PSUD}), while two regimes are robustly found for the majority of languages when specific sentence lengths are considered (\autoref{fig:best_model}). However, while the picture is clear and consistent in the first case, discussion on sentences of specific lengths requires further elaboration. The shape of the distribution depends on the length of the sequence (\autoref{fig:best_model}), which is expected by the relation between DDm and sentence length. Processing short distances implies lower cognitive effort and robust statistical evidence suggests that DDm might irrelevant or be canceled out by other word order principles in short sequences \parencite{ref:Ferrer-i-Cancho2021, Omega, Ferrer2023b}. Then, the varying intensity of the pressure for minimization yields different distributions in different areas of the sentence length domain, which we characterized with the following (potentially overlapping) regions (\autoref{fig:best_model}, \autoref{fig:PUD_omega} and \autoref{fig:PSUD_omega}).

\begin{itemize}
    \item \textit{Random linear arrangement.} In short sentences (approximately $n \leq 6$) DDm might be neglectable or weak enough to be surpassed by other word order principles \parencite{ref:Ferrer-i-Cancho2021, Omega, Ferrer2023b}, resulting in Model 0 (green tiles in \autoref{fig:best_model}, \autoref{fig:PUD_omega} and \autoref{fig:PSUD_omega}) sometimes being the best one to describe the distribution. Where it is not Model 0, a model with a single regime is the best one.
    \item \textit{Single chunk.} Up to roughly 13 words the best model is mainly one of 1, 2, or 5 (yellow and pink tiles in  \autoref{fig:best_model}, \autoref{fig:PUD_omega} and \autoref{fig:PSUD_omega}) in most languages. This possibly indicates that the sentence can be processed as a single chunk when the number of words is small enough, and dependencies must be highly local to allow for this.
    \item \textit{Two regimes.} The bulk of the sentence length domain is characterized by the presence of two-regime models (red and blue tiles in \autoref{fig:best_model}, \autoref{fig:PUD_omega} and \autoref{fig:PSUD_omega}). In these longer sentences the burden on STM becomes heavier, and two regimes might emerge from the breaking down of the sentence into chunks. After a very steep decrease in probability, a long dependency becomes more likely in order to link a chunk to the previous one.
    \item 
    \textit{No consistent pattern}.
    For long (and rare) sentences no clear pattern appears, as the scarcity of examples for large sentence lengths introduces variability in the estimation of the best model.
\end{itemize}
\iftoggle{Sonia}{
\textcolor{red}{Sonia: I cannot follow following the argument in the following paragraph that is hidden as a commentary in the .tex source; I would remove it but I would like you to check it again before we take the ultimate decision}
}

\subsubsection{On power laws}


When mixing sentences of distinct length, the best model is always a two regime model (\autoref{tab:best_PUD_PSUD}).
Across sentence lengths, the majority of languages have a model with two regimes as the most frequent best one, and a few languages in both collections show a power-law behavior (\autoref{tab:best_PUD_PSUD_fixed}). 
Nevertheless, setting a rather 
high
representativeness threshold dramatically reduces evidence for single-regime power-law, especially in PUD (\autoref{fig:threshold}). This is for instance the case with Chinese in both collections. In spite of this, for Arabic, Indonesian, and Thai the most frequent best model is robustly Model 5 when the SUD annotation style is used. 

Although Chinese has been argued to follow a single-regime power law \parencite{ref:HaitaoLiu}, our findings indicate that Chinese is better fitted by a two-regime model with an initial power-law regime (Model 6 or 7) when mixing sentences of any length (\autoref{tab:best_PUD_PSUD}). However, if the representativeness threshold is set to a low value (\autoref{fig:threshold}), a single-regime power law (Model 5) can be retrieved, but such a low threshold casts doubts on the statistical strength of the best model when mixing sentences of distinct length. In contrast, the claim of a power law for Chinese is supported clearly for sentences of fixed length, where Model 5 is the most frequent best model across sentence lengths (\autoref{tab:best_PUD_PSUD_fixed}). 

Overall, two exponential regimes are the most common distribution for both mixed and fixed sentence lengths. However, what our analysis also proposes is that power laws can well describe the distribution in the first regime for some languages (mainly non Indo-European) when sentence lengths are mixed, as well as the distribution for specific sentence lengths for a small subset of them. Importantly, power-laws can also arise from undersampling, as highlighted by our representativeness analysis (\autoref{fig:threshold}). In previous research it has been argued that power-laws could emerge from mixing sentence lengths in which distances are distributed following an exponential curve \parencite{Stumpf10022012,riskofmixing}. Our research invalidates this argument (at least in the scope of our sample of languages), and identifies instances of another sort of mixing: for Arabic, Indonesian, and Thai in PSUD, mixing sentence lengths that are individually power-law distributed results in a distribution with two regimes with a power-law in the 1st regime, suggesting that further investigation is required in this direction.

\subsubsection{Tail variability}

Plots of the best model against the real data allows one to visually assess the quality of its fit to the data (\autoref{fig:PUD_best_fitted} for PUD and \autoref{fig:PSUD_best_fitted} for PSUD). The best models are able to very well capture the shape of the bulk of the distribution and the initial bending in all languages. However, they are not always able to fully capture the variability along the tail of the distribution. To begin with, noise naturally emerges for longer distances, which belong to rare long sentences. As we explained above, there are lengths for which only one sentence is observed. Taking this into account, the deviation from the best model could suggest the possible presence of an unveiled pattern for some languages. We hypothesize the existence of more than one break-point, implying 
incremental executions of a ``chunk-and-pass'' mechanism \parencite{Christiansen2016a}. 

However, introducing more regimes would greatly increase both the complexity of estimation (maximum likelihood estimation already requires putting particular care in the estimation of 3/4 parameters, see section \ref{sec:Methods}), and the risk of overfitting the data. Thus, a thorough and rigorous methodology would need to be employed for such modelling, which should be the subject of future research.

\subsection{The homogeneity of the break-point}

The break-point values we estimated are largely homogeneous across languages, and average values of 5 (PUD) and 4 (PSUD) words, with small variation. These values are consistent with the literature on limitations of short term memory: in no language $d^*$ exceeds the ``magical number'' 7 \parencite{Miller1956}, and the bulk of the values is centered at $4 \pm 1$, which is generally recognized to be the working memory limitation on a wide range of tasks \parencite{Cowan2001}. 

Nevertheless, some variability can still be observed, especially among the break-points of sentences of different lengths within a language. In fact, an implicit assumption of $H_2$ is that the value estimated globally for a given language is a reliable approximation of the constraint acting at the sentence level, and this can be verified by looking at the break-points estimated for each given length. We find that for languages in which $H_1$ holds (two exponential regimes), the distribution of $d^*$ across sentence lengths is very narrow, and centered around the global value of $d^*$. The break-points estimated in Models 6-7 are more variable, but they still vary in a rather small range compared to the range of variation of the actual sentences (\autoref{tab:PUD_summary} and \autoref{tab:PSUD_summary}).

The average length in words of simple declarative sentences is 3.7 (from 2.6 in Turkish up to 5.4 in Mandarin)
\parencite{Fenk-Oczlon2021a}.\footnote{The data can be found in the Supplementary Material (Sheet 1)}. We believe that this variability in the size clauses is captured by our breakpoint (\autoref{fig:dstar_violinplot}) but this issue should be the subject of future research with a linguistic or cognitive focus.

\subsection{Patterns in probability decay across regimes}

Given the large applicability of the two-regime models, 
we take closer look to the speed of probability decay. 
The slopes observed across languages are quite narrowly distributed around the same values (\autoref{fig:mixed_n_slopes}). It is interesting to notice that while the first slope is significantly larger in PSUD, $q_2$ shows little variation in the two collections. This suggests that, depending on annotation style, the distribution of the dependencies within word chunks will change, but beyond word chunks, the chunking mechanism follows a similar structure. Another interesting pattern concerns the steepness of the first regime with respect to the second one. When mixing sentences of different lengths the first regime is always steeper than the second one (\autoref{fig:mixed_n_slopes}) and this is virtually always the case even when considering specific sentence lengths, with a very few exceptions in short sentences (\autoref{fig:fixed_n_slopes}). This provides additional support for the ``chunk-and-pass'' paradigm \parencite{Christiansen2016a}. An explanation for that pattern could be that, when memory limits are approached in long enough sentences, the current chunk needs to be closed, and a new longer dependency becomes more likely in order to link the forthcoming chunk (thus reducing the speed of probability decay). 
\iftoggle{Sonia}{
\textcolor{red}{Sonia: I do not understand the following arguments (text up to the next subsection). We may have to remove them for simplicity or given the primary audience of the journal; another reason for removing or smoothing is that if refers to BIC, which is very technical and the previous sections do not recur to specific concepts such as BIC/AIC.}
However, there might be an opposite mechanism that may explain the exceptions, namely, when the sentence is short enough to be processed as a single chunk, once the constraint on STM is reached, pressure for minimization increases, thus also increasing the slope of the curve in the second regime. There may be a second mechanism for these exceptions given our findings with model selection on artificial data \textcolor{red}(now these results are located in an appendix???)} The two regimes (and in particular Model 4) may be found even if the real distribution is Model 0, given their similar BIC scores (\autoref{fig:ms_artificial_BICs}). However, Model 4 could only mimic a linear curve (Model 0) if the second regime was steeper than the first one.
}

\subsection{The best model versus the optimality of syntactic dependency distances}


In section \ref{subsec:two_regimes}, we have described how the shape of the distribution varies depending on sentence length. Here, we aim to understand the interplay with  different degrees of pressure for DDm for long versus short sentences. Previous research has pointed out at how $\left <\Omega \right>$ is smaller in short sentences, likely due to DDm being neglected or canceled out by other word order principles \parencite{ref:Ferrer-i-Cancho2021, Omega, Ferrer2023b}. We provide additional evidence for this phenomenon by unravelling a direct correspondence between sentences where $\left <\Omega \right>$ is close to 0, and those in which the best model is Model 0 (\autoref{tab:omega_0}). Moreover, we observe a relation between the intensity of DDm and the best model for the distribution. Namely, as pressure for minimization increases with sentence length, the best model changes (\autoref{fig:PUD_omega} and \autoref{fig:PSUD_omega}). While correlation does not imply causation, it is crucial to understand that both the pressure for DDm and the best model for the distribution of syntactic dependency distances are not homogeneous through sentence length. Thus, distances belonging to sentences of different length are subject to different pressures, and this should be taken into account when trying to model the distribution. In particular, these different levels of pressure could yield different mechanisms. Indeed, the more complex distributions -- those with two regimes -- tend to emerge for long enough sentence length, when the pressure for DDm is stronger, likely calling for a structured processing mechanism.


\subsection{The effect of annotation style}


So far we have observed commonalities and differences between PUD and PSUD.
Overall, the main qualitative results are robust to annotation style, supporting the soundness of the observed patterns, but some differences emerge. The discussion on the origins of such differences is open, and is connected to the fundamental question of whether an annotation style is a more accurate representation of our brain's functioning or the linguistic processing than the other, or whether different styles simply mirror different aspects of this functioning or processing. While providing a rather descriptive account of such differences, we partly attempt to address this question.

\subsubsection{The shape of the distribution}

The first main point concerns the very high consistency in the best estimated models (\autoref{fig:dstar_violinplot}). However,  there are a few exceptions, which we classified in two types: differences in right truncation, and in the distribution in the first regime. The latter is clearly of greater interest and it concerns two languages, Japanese and Indonesian, both having Models 3-4 as the best model in PUD, and Model 7 in PSUD, but showing a very different behaviour. For  Japanese, the best models estimated on specific sentence lengths and by mixing all sentence lengths are highly consistent within each collection, and in both cases the break-point value is $d^*=6$. This suggests a real difference in probability decay within a chunk depending on the chosen annotation guidelines, but also conveys the concreteness of the quantified limit on memory for such language. On the other hand, for Indonesian we find mixed evidence, both in terms of estimated break-point, which goes from $d^*=3$ in PUD to $d^*=7$ in PSUD, and in terms of best model for fixed sentence lengths (which is consistently a one-regime power-law in PSUD). In fact, this takes us to one of the main differences between annotation styles (\autoref{fig:threshold}): while in PUD the only language showing some evidence for a single power-law regime for fixed sentence lengths is Arabic, in PSUD we have three languages strongly supporting the reality of such distribution. For Arabic, Indonesian, and Thai, the two regimes observed for mixed sentence lengths contradict what is found when sentence lengths are analysed in isolation. This seems to reflect Simpson's paradox, a phenomenon according to which a statistical trend disappears when single groups are considered, and suggests that there is some variability left to explain.

\subsubsection{The break-point}

We have seen in \autoref{fig:dstar_violinplot} how the break-points estimated in both collections cover the same portion of domain, ranging from 3 to 7. 
However, while in PUD there is no settling around a particular value, in PSUD $d^*$ is nearly uniform at $d^*=4$, especially within Models 3-4. This raises the following questions: is this regularity given by chance? Or does it mirror a better ability of SUD to capture syntactic relations as formed by our minds? Given that -- besides individual differences -- the overall structure of the brain is assumed to be the same for all humans, the constraint on memory is expected to be uniform across languages (hence the motivation for $H_2$). Thus, one could speculate that SUD annotation style is actually more capable of unveiling this uniformity, that is assumed to exist.

\subsubsection{Dependency distance minimization}


SUD guidelines have been found to lead to shorter dependency distances \parencite{Osborne2019a,Yan2021a,Omega}. When dependency distances are conveniently normalized with respect to the gap between the random baseline and the minimum baseline, SUD reflects distances that are closer to optimality \parencite{Omega}. Such ability of SUD to reflect dependency distance minimization of effects is confirmed by our findings. 
In fact, despite predicting a power-law decay in the first regime for two more languages compared to PUD, $q_1$ is significantly higher in PSUD (\autoref{fig:mixed_n_slopes}). This entails a faster decay in probability within the chunk, related to the predominance of short local dependencies in PSUD. Moreover, the values of $\Omega$ computed in the PSUD collection are generally larger (tiles  in \autoref{fig:PSUD_omega} (b) are darker than in  \autoref{fig:PUD_omega} (b)), confirming a stronger degree of optimization of dependency distances in the SUD framework \parencite{Omega}.

\section{Conclusion}
\label{sec:Conclusions}
Two decades after 
the first observations on the peculiar shape of the distribution of syntactic dependency distances
\parencite{ref:Ferrer-i-Cancho2004}, some new light has been shed. A crucial finding is that the probability of observing a dependency -- independently of the length of the sentence it belongs to -- is best described by a double-regime model. Furthermore, the finding also holds at a finer-grained level, distinctively considering each sentence length. In this setting, for the great majority of languages a double-regime model is the most frequent one, while the few remaining languages show a power-law decay as the most frequent, partly in accordance with what has been found concerning a Chinese treebank, where however sentences of mixed lengths were analysed \parencite{ref:HaitaoLiu}. Furthermore, the break-point between the two regimes estimated globally for each language varies in a small range ($3 \leq d^* \leq 7$), which becomes even narrower when only languages in which $H_1$ holds are considered. In fact, $H_2$ seems to be related to the probability distribution observed in the first regime, leading to the identification of a group of languages where probability follows a two-regime exponential decay ($H_1$), and within which the break-point is very similar ($H_2$). This group is mainly populated by Indo-European languages. However, languages from this family are over-represented in our sample, and other interesting patterns could emerge if a larger group of languages from other families where analysed. These considerations hold independently of annotation style, but it has not escaped our attention that in PSUD values of $d^*$ for such group are almost uniform at 4, a widely accepted quantification of the constraint on short term memory \parencite{Cowan2001}. This could, in our opinion, reflect a higher sensitivity of SUD annotation style to the way in which our minds create and process language, bringing to light a ``universal'' constraint which is not language dependent. Another general pattern emerged is the relation between the speeds of the decays, whereas probability in the first regime is always faster than in the second one. As already pointed out, this result may look paradoxical: if cognitive pressure induces a decay in probability as syntactic dependency distance increases, why does such a decay slows down beyond the breakpoint? \parencite{Ferrer2017d}?
In the framework of language processing, these findings provide strong support for the ``chunk-and-pass'' mechanism \parencite{Christiansen2016a}. In fact, the presence of these two different regimes could actually mirror the two different speeds at which probability decays within a chunk and beyond it. In physical terms, the true units of measurement of distance may change: within the word chunk the unit of distance are words whereas, beyond the word chunk, the actual distance may be chunks in the hidden space of incremental processing of the sentence. The breakpoint and the slow down after the breakpoint may arise because we have imposed the use of words as unit of measurement independently of the stage of syntactic parsing.  In our view, this appears to be the most reasonable and pertinent explanation for the observed systematic decrease in the strength of DDm, but we do not exclude that other explanations could as well be plausible. Future work could further investigate the distribution in the second regime, exploring different combinations of exponential and power-law decay. Then, the possible presence of more than one break-point could be explored. 
Importantly, to understand the extent to which the observed phenomena can be considered universal, the same analysis shall be performed on a wider set of languages. 

\section*{Acknowledgments}

We are grateful to Jan Andres for helpful comments. 
We have benefited from discussions with G. Fenk-Oczlon and the contents of the talk that she gave at the 16th International Cognitive Linguistics Conference (August 2023), ``Working memory constraints: Implications for efficient coding of messages''. They helped us in terms of presenting STM for a general audience and to find a linguistic interpretation to the breakpoint.
SP is funded by the grant ``Thesis abroad 2021/2022'' from the University of Milan.
RFC is supported by a recognition 2021SGR-Cat (01266 LQMC) from AGAUR (Generalitat de Catalunya). SP and RFC are supported by the grants AGRUPS-2022, AGRUPS-2023 and AGRUPS-2024 from Universitat Politècnica de Catalunya.


\printbibliography



\appendix

\section*{Appendices}

\section{Model derivation}
\label{sec:AppendixA}
Here we detail the mathematical derivation of the non-standard models in Section \ref{sec:Models}.

\paragraph{Model 0.1}

We consider a general model for sentences of varying length, defined as
\begin{equation*}
p(d) = \sum_{n=min(n)}^{max(n)} p(d | n) \: p(n),
\end{equation*}  
where $p(d | n)$ is the conditional probability of $d$ given that the sentence length has $n$ words, $p(n)$ is the proportion of sentences having length $n$, and $\min(n)$ and $\max(n)$ are the minimum and maximum observed values of $n$ in the sample. By definition, $p(d | n)$ satisfies two conditions, i.e. $p(d | n) = 0$ when $d \notin [1, n)$
and 
\begin{equation*}
\sum_{d=1}^{n - 1} p(d | n) = 1.    
\end{equation*}
Thanks to these two conditions, it is easy to see that $p(d)$ is properly normalized, that is 
\begin{eqnarray*}
\sum_{d=1}^{max(n) - 1} p(d) & = & \sum_{d=1}^{max(n) - 1} \sum_{n=min(n)}^{max(n)} p(d | n) \: p(n) \\
                             & = & \sum_{n=min(n)}^{max(n)} p(n) \sum_{d=1}^{n - 1} p(d | n) \\
                             & = & 1
\end{eqnarray*}
By setting $p(d | n)$ according to the null hypothesis of a random shuffling of the words of a sentence of $n$ words (\ref{eq:model0}), which satisfies the two conditions above, we obtain
\begin{equation*}
p(d) = \sum_{n=min(n)}^{max(n)} \frac{n-d}{\binom{n} {2}} \: p(n).
\end{equation*}

\paragraph{Model 2}
We define the cumulative distribution of Model $1$ as 
\begin{equation*}
    P_1(d) = \sum_{d'=1}^{d} p_1(d').
\end{equation*}
where $p_1(d)$ is defined as in \ref{eq:model1}.
Model 2 is derived via renormalization of Model 1 after right-truncation, that is 
\begin{equation*}
    p_2(d) = \frac{p_1(d)}{P_1(d_{max})},
\end{equation*}
where
\begin{eqnarray*}
    P_1(d_{max}) & = & \sum_{d=1}^{d_{max}} q(1-q)^{d-1} \\
    & = & 1 - (1-q)^{d_{max}}.
\end{eqnarray*}
Hence
\begin{equation*}
    p_2(d) = \frac{q(1-q)^{d-1}}{1 - (1-q)^{d_{max}}}.
\end{equation*}

\paragraph{Double-regime models}

Now we use $p_1(d)$ to refer to the definition of $p(d)$ for $d \leq d^*$ and 
$p_2(d)$ to refer to the definition of $p(d)$ for $d \geq d^*$. The definition of Models 3, 4, 6, 7 follows the template
\begin{equation*}
p(d) = \left\{
          \begin{array}{ll} 
              p_1(d) = c_1 f_1(d) & \mbox{if~} d \leq d^* \\ 
              p_2(d) = c_2 f_2(d) & \mbox{if~} d^* \leq d \leq d_{max}, \\ 
          \end{array}
        \right.
\end{equation*}
For models 3 and 6, one simply sets $d_{max}$ to $\infty$.
Thus, the assumption $p_1(d) = p_2(d)$ yields 
\begin{equation*}
c_2 = \tau c_1
\end{equation*}
with
\begin{equation*}
\tau = \frac{f_1(d)}{f_2(d)}.
\end{equation*}
Recalling the definitions of the models (\autoref{tab:models}), it is easy to see that, for models 3 and 4, 
\begin{equation*}
  \tau = \frac{(1-q_1)^{d^*-1}}{(1-q_2)^{d^*-1}}.
\end{equation*}
whereas for models 6 and 7, 
\begin{equation*} 
  \tau = \frac{d^{*^{-\gamma}}}{(1-q)^{d^*-1}}.
\end{equation*}

Let us derive the normalization factor $c_1$ for Models 3, 4, 6, 7 with the help of 
\begin{eqnarray*}
S_1 & = & \sum_{d=1}^{d^*} f_1(d) \\
S_2 & = & \sum_{d=d^*}^{d_{max}} f_2(d).
\end{eqnarray*}
The normalization condition 
\begin{equation*}
\sum_{d=1}^{d_{max}} p(d) = 1
\end{equation*}
yields
\begin{equation}
\label{eq:c1_models}
c_1 = \frac{1}{S_1 + \tau S_2}.
\end{equation}

For Models 3 and 4, $S_1$ is 
\begin{equation*}
    S_1 = \sum_{d'=0}^{d^*-1} (1-q_1)^{d'} = \frac{1-(1-q_1)^{d^*}}{q_1}.
\end{equation*}
$S_2$ depends on the truncation point. For Model 3, the assumption $q>0$ (thus $\lim_{d_{max}\to\infty} (1-q)^{d_{max}} = 0$) produces 
\begin{eqnarray}
    S_2 & = & \sum_{d'=d^*}^{\infty} (1-q_2)^{d'} \nonumber \\
    (1-q_2)S_2 & = & S_2 - (1-q_2)^{d^*} + (1-q_2)^\infty \nonumber  \\
    S_2 & = & \frac{(1-q_2)^{d^*}}{q_2} \label{eq:S2_models3-6}
\end{eqnarray}
By substituting $S_1$, $S_2$ and $\tau$ in \ref{eq:c1_models}, 
$c_1$ for Model 3 becomes 
\begin{equation*}
    c_1 = \frac{q_1q_2}{q_2 + (1-q_1)^{d^*-1}(q_1 - q_2)}
\end{equation*}
after some algebra. 

In Model 4, probabilities are restricted up to $d_{max}$, thus
\begin{equation}
\label{eq:S2_models4-7}
    S_2 = \sum_{d'=d^*}^{d_{max}-1} (1-q_2)^{d'} = \frac{(1-q_2)^{d^*} - (1-q_2)^{d_{max}}}{q_2}.
\end{equation}
Again, plugging $S_1$, $S_2$, and $\tau$ into \ref{eq:c1_models} produces $c_1$ for Model 4, that is
\begin{equation*}
    c_1 = \frac{q_1q_2}{q_2 + (1-q_1)^{d^*-1}(q_1 - q_2 - q_1(1-q_2)^{d_{max} - d^*+1} )}
\end{equation*}
after some algebra. 

For the second pair of double-regime models (Models 6 and 7), combining a zeta and a geometric distribution, $S_1$ is 
\begin{equation*}
    S_1 = \sum_{d=1}^{d^*} d^{-\gamma} = H(d^*,\gamma),
\end{equation*}
while the second regime is shared with Models 3-4, so that $S_2$ corresponds to \ref{eq:S2_models3-6} for Model 6 and to \ref{eq:S2_models4-7} for Model 7. Then, the normalization factors are obtained again through \ref{eq:c1_models}, so that for Model 6
\begin{equation*}
    c_1 = \frac{q}{q H(d^*,\gamma) + d^{*^{-\gamma}} (1-q)},
\end{equation*}
while for Model 7 
\begin{equation*}
    c_1 = \frac{q}{q H(d^*,\gamma) + d^{*^{-\gamma}} (1-q - (1-q)^{d_{max}-d^*+1})}
\end{equation*}
after some algebra. 

\newpage

\section{Log-likelihood functions}
\label{sec:Appendix_log_likelihood}
 
In our setting, the log-likelihood of a model is 
\begin{equation*}
\pazocal{L} = \log \prod_{i=1}^N p(d_i) = \sum_{i=1}^{N} \log p(d_i) = \sum_{d=1}^{max(d)} f(d), \log p(d).
\end{equation*}
Next we derive the log-likelihood functions for each model with the help of \autoref{tab:models}.

For Model 0.0, where $d_{max}$ is the only free parameter, we have
\begin{eqnarray*}
\pazocal{L}  & = & \sum_{d=1}^{max(d)}  f(d) \log \left(\frac{2(d_{max}+1-d)}{d_{max}(d_{max}+1)} \right) \\
 & = & \sum_{d=1}^{max(d)}  f(d) \left[ \log \left(\frac{2}{d_{max}(d_{max}+1)} \right) + \log (d_{max}+1-d) \right] \\
 & = & N \log \left(\frac{2}{d_{max}(d_{max}+1)} \right) +  W, 
\end{eqnarray*}
where 
\begin{eqnarray*}
N & = & \sum_{d=1}^{max(d)}  f(d) \\
W & = & \sum_{d=1}^{max(d)}  f(d) \log (n-d).
\end{eqnarray*}
For Model 0.1, in which the observed sentence lengths are supplied and there is no free parameter, we have
\iftoggle{Sonia}{
\textcolor{red}{Sonia: I think that the following derivation is incorrect or the actual definition of model 0.1 is unclear to me}
}{}
\begin{eqnarray*}
\pazocal{L}  & = & \sum_{n=min(n)}^{max(n)} \sum_{d=1}^{max(d)}  f(d) \log \frac{2(n-d)}{n(n-1)} \\
 & = & \sum_{n=min(n)}^{max(n)} \sum_{d=1}^{max(d)}  f(d) \left[ \log \frac{2}{n(n-1)}  + \log (n-d) \right] \\
 & = & \sum_{n=min(n)}^{max(n)} \left[ N_n \log \frac{2}{n(n-1)} +  W_n  \right]
\end{eqnarray*}
where 
\begin{eqnarray*}
W_n & = &\sum_{d=1}^{max(d)} f(d) \log (n-d) \\  
N_n & = & \sum_{d=1}^{max(d)} f(d)
\end{eqnarray*}
in sentences of length $n$. 
\iftoggle{Sonia}{
\textcolor{red}{Sonia: What does the last phrase mean? Does it mean that the summations are restricted to sentences of length $n$? If yes the formulae need to be improved.}
}{}
For the \textit{geometric models}, we start from the derivation of the right-truncated version, namely Model 2
\begin{eqnarray*}
 \pazocal{L}  & = & \sum_{d=1}^{max(d)}  f(d) \log \frac{q(1-q)^{d-1}}{1-(1-q)^{d_{max}}} \\
 & = & \sum_{d=1}^{max(d)}  f(d) \left[ \log \frac{q}{1-(1-q)^{d_{max}}}  +  (d-1) \log (1-q) \right] \\
 & = & N \log \frac{q}{1-(1-q)^{d_{max}}} + (M-N) \log(1-q),
\end{eqnarray*}
where $M = \sum{d=1}^{max(d)} f(d) \: d$. Then, the log-likelihood function of Model 1 as a particular case of that of Model 2 in which $d_{max} = \infty$, i.e.
 \begin{equation*}
  \pazocal{L} = N \log q + (M - N) \log (1-q)
 \end{equation*}
since $q>0$ and thus $\lim_{d_{max}\to\infty} (1-q)^{d_{max}} = 0$.
For the \textit{two-regime geometric models}, we start from the log-likelihood of Model 4, i.e.
\begin{eqnarray*}
 \pazocal{L}  & = & \sum_{d=1}^{d^*}  f(d) \log \left[ c_1(1-q_1)^{d-1}\right] + \sum_{d=d^*+1}^{max(d)}  f(d) \log \left[ c_2(1-q_2)^{d-1}\right] \\
 & = & \sum_{d=1}^{d^*}  f(d) \left[ \log c_1 + (d-1)\log(1-q_1) \right] + \sum_{d=d^*+1}^{max(d)}  f(d) \left[ \log c_2 + (d-1)\log(1-q_2) \right] \\
 & = & N^*\log c_1  + (M^* - N^*) \log(1-q_1) + (N-N^*)\log c_2 + \\
 &   & (M - M^* - N + N^*)\log(1-q_2) \\
 & = & N^* \log c_1  + (N-N^*) \log c_2 + (M^*- N^*) \log \frac{1-q_1}{1-q_2} + (M-N)\log(1-q_2)
\end{eqnarray*}
where 
\begin{eqnarray*}
 M^* & = & \sum_{d=1}^{d^*} f(d) \: d \\
 N^* & = & \sum_{d=1}^{d^*} f(d),
\end{eqnarray*} 
while $c_1$ and $c_2$ are defined as explained in Section \ref{sec:Models} for Model 3 and 4.
Thus, the log-likelihood functions of Model 3 and Model 4 only differ in the computation of $c_1$ and $c_2$. For the \textit{right truncated power-law} distribution, namely Model 5,
\begin{eqnarray*}
 \pazocal{L}  & = & \sum_{d=1}^{max(d)}  f(d) \log  \frac{d^{-\gamma}}{H(d_{max},\gamma)} \\
 & = & \sum_{d=1}^{max(d)}  f(d) \left[ -\gamma \log d - \log H(d_{max},\gamma) \right] \\
 & = & -\gamma M' - N \log H(d_{max},\gamma),
\end{eqnarray*}
where $ M' = \sum_{d=1}^{max(d)}  f(d) \log (d)$. Finally, for Models 6 and 7, we start from the derivation of Model 7, 
\begin{eqnarray*}
 \pazocal{L}  & = & \sum_{d=1}^{d^*}  f(d) \log (c_1 d^{-\gamma}) + \sum_{d=d^*+1}^{max(d)}  f(d) \log \left[c_2(1-q)^{d-1}\right] \\
 & = & \sum_{d=1}^{d^*}  f(d) \left[ \log c_1 - \gamma \log(d) \right] + \sum_{d=d^*+1}^{max(d)}  f(d) \left[ \log c_2 + (d-1)\log(1-q) \right] \\
 & = & N^*\log c_1  - \gamma M'^* + (N-N^*)\log c_2 + (M - M^* - N + N^*)\log(1-q),
\end{eqnarray*}
while $c_1$ and $c_2$ are defined as explained in Section \ref{sec:Models} for Model 6 and 7.

\begin{sidewaystable}[ht]
\caption{\label{tab:loglikelihoods} The log-likelihood $\pazocal{L}$ for each of the probability mass functions. $K$ is the number of free parameters, $N$ is the sample size, $M$ is the sum of distances weighted by frequency, i.e. $M = \sum_{i=1}^{max(d)} f(d_i) d_i$, $M^*$ is the same sum up to $d^*$, i.e. $M^* = \sum_{i=1}^{d^*} f(d_i)d_i$, $M'$ is the sum of log distances weighted by frequency, i.e. $M' = \sum_{i=1}^{max(d)} f(d_i) log(d_i)$, and $W$ is such that $W = \sum_{i=1}^{max(d)} f(d_i) log( d_{max} + 1 - d_i)$. $N^*$ is the sum of distance frequencies up to $d^*$, i.e.  $N^* = \sum_{i=1}^{d^*} f(d_i)$, and $M'^*$ is $M'$ up to $d^*$, i.e. $M'^* = \sum_{i=1}^{d^*} f(d_i) log(d_i)$. For model 3, $c_1$ id defined in \ref{eq:c1_model3}; for model 4, $c_1$ is defined in \ref{eq:c1_model4}; for models 3 and 4, $c_2 = \tau c_1$ with $\tau$ defined as in \ref{eq:tau_models3-4}.
For model 6, $c_1$ id defined in \ref{eq:c1_model6}; for model 7, $c_1$ is defined in \ref{eq:c1_model7}; for models 6 and 7, $c_2 = \tau c_1$ with $\tau$ defined as in \ref{eq:tau_models6-7}.
Finally, $W_n = \sum_{i=1}^{max(d)} f(d_i) \log (n-d_i))$ and $N_n = \sum_{i=1}^{max(d)} f(d_i) $ in sentences of length $n$. 
\iftoggle{Sonia}{\textcolor{red}{Sonia 1: What does the last phrase mean? Does it mean that the summations are restricted to sentences of length $n$? If yes the formulae need to be improved.}
\textcolor{red}{Sonia 2: In this table and in the text in this appendix I replaced $r_1$ and $r_2$ (which looked outdated notation to me), by $c_1$ and $c_2$ respectively. Please, ensure that my change is correct. }
}{}
\iftoggle{Sonia}{\textcolor{red}{Now there are rows for Model 0.1 and Model 0.1. There should be a row for Model 0 if it is likelihood is computed somewhere.}}{}
}
\begin{tabular}{@{}*{4}{l}}
\toprule
 \textbf{Model} & \textbf{Function} & $K$ & $\pazocal{L}$ \\ 
\midrule
 0.0 & Null model   & 1 & $ N \log(\frac{2}{d_{max}(d_{max}+1)}) + W$ \\
0.1 & Extended Null model   & 0 & $\sum_{n=min(n)}^{max(n)} [ N_n \log \left(\frac{2}{n(n-1)} \right) +  W_n  ]$ \\
 1 & Geometric    & 1 & $ N \log q + (M - N) \log (1-q)$ \\
 2 & Right-truncated geometric   & 2 & $N \log \left(\frac{q}{1-(1-q)^{d_{max}}} \right) + (M-N) \log(1-q)$ \\
 3 & Two-regime geometric & 3 & $N^* \log c_1 + (N-N^*) \log c_2 + (M^*- N^*) \log \left( \frac{1-q_1}{1-q_2} \right) + (M-N)\log(1-q_2)$\\
 4 & Two-regime right-truncated geometric & 4 & 
 $N^* \log c_1 + (N-N^*) \log c_2 + (M^*- N^*) \log \left( \frac{1-q_1}{1-q_2} \right) + (M-N)\log(1-q_2)$\\
 5 & Right-truncated zeta distribution   & 2 & 
 $ -\gamma M' - N \log(H(d_{max},\gamma))$ \\ 
 6 & Two-regime zeta-geometric & 3 & $N^* log(c_1) - \gamma M'^* + (N-N^*) log(c_2) + (M-M^* -N + N^*)log(1-q)$ \\
 7 & Two-regime right-truncated zeta-geometric & 4 & $N^* log(c_1) - \gamma M'^* + (N-N^*) log(c_2) + (M-M^* -N + N^*)log(1-q)$ \\
\bottomrule
\end{tabular}
\end{sidewaystable}

\section{Model selection validation}
\label{sec:AppendixAB}

\subsection{Artificial data generation}

In the following, let $p_x(d)$ be the probability of $d$ according to Model x. The parameter values used to generate each model are reported in \autoref{tab:artificial_params_generation}, while sample size is $N = 10^4$ for each model. For right-truncated models sentence length is set to $n = 20$, 
and the maximum distance is set to 
$d_{max}=19$. 
\iftoggle{Sonia}{\textcolor{red}{Model 0 here could be Model 0 or Model 0.0}}{Then Model 0.0 is equivalent to Model 0 with $n = 20$.}
We choose $\gamma=1.6$ because it has been obtained from fitting a right-truncated Zeta distribution to a Chinese treebank \parencite{ref:HaitaoLiu}.
 
\begin{table}[H]
\caption{\label{tab:artificial_params_generation} Parameter values used to generate artificial samples. \iftoggle{Sonia}{\textcolor{red}{Model 0 here could be Model 0 or Model 0.0}}{Here Model 0 is the same as Model 0.0.}
}
\centering
\begin{tabular}{@{}*{7}{l}}
\toprule
\textbf{Model} & $d_{max}$ & $q$ & $q_1$ & $q_2$ & $d^*$ & $\gamma$\\
\midrule
   \textit{0} & 19 & - & - & - & - & - \\
   \textit{1} & - & 0.2 & - & - & - & - \\
   \textit{2} & 19 & 0.2 & - & - & - & - \\
   \textit{3} & - & - & 0.5 & 0.1 & 4 & - \\
   \textit{4} & 19 & - & 0.5 & 0.1 & 4 & - \\
   \textit{5} & 19 & - & - & - & - & 1.6 \\ 
   \textit{6} & - & 0.2 & - & - & 4 & 1.6 \\
   \textit{7} & 19 & 0.2 & - & - & 4 & 1.6 \\
\bottomrule
\end{tabular}
\end{table}

\paragraph{Models 1 and 2}

For the geometric distribution and its right-truncated version, namely Model 1 and Model 2, we use Dagpunar's fast inversion method \parencite{dagpunar1988principles}. For Model 1, a random distance $d$ is obtained by producing a random uniform deviate $x$ and then calculating 
\begin{equation*}
    d = 1 + \left\lfloor\frac{\log x}{\lambda}\right\rfloor,
\end{equation*}
where $\lambda = \log (1-q)$, and $q$ is the parameter of the desired geometric distribution. 
For Model 2, a value of $d$ is produced until $d \leq d_{max}$. 

\paragraph{Model 5}
For Model 5, we employed the algorithm proposed by Devroye to efficiently generate a random deviate from a zeta distribution \parencite{ref:LucDevroye}, adapting it to allow for right-truncation. The algorithm is called one or more times until a value of $d$ such that $d \leq d_{max}$ is obtained. 


\paragraph{Model 0 and two-regime models}

For the sake of simplicity, random samples of Model 0 and of the two-regime models, namely Models 3, 4, 6, and 7, are generated using a tabular inversion method \parencite{muller1958inverse,ref:LucDevroye}. 
\iftoggle{Sonia}{
\textcolor{red}{Sonia: notice that I have introduced $\delta$ to avoid ambiguous uses of $d_{max}$. }
} {}
This method generates artificial distances in a pre-specified range, namely $d \in [1,\delta]$. Thus, in order to simulate Models 3 and 6 -- which do not have a right-truncation -- we set $\delta=10^6$ to ensure that $p(d) \approx 0$ for $d \geq \delta$, while for Models 0, 4 and 7 we have $\delta = d_{max}=19$. For simplicity, the method is implemented through binary search. Hence, a random deviate is produced in time $O(\log \delta)$.

\subsection{Results}

\begin{figure}[H]
\centering
    \includegraphics[width = 0.85\textwidth]{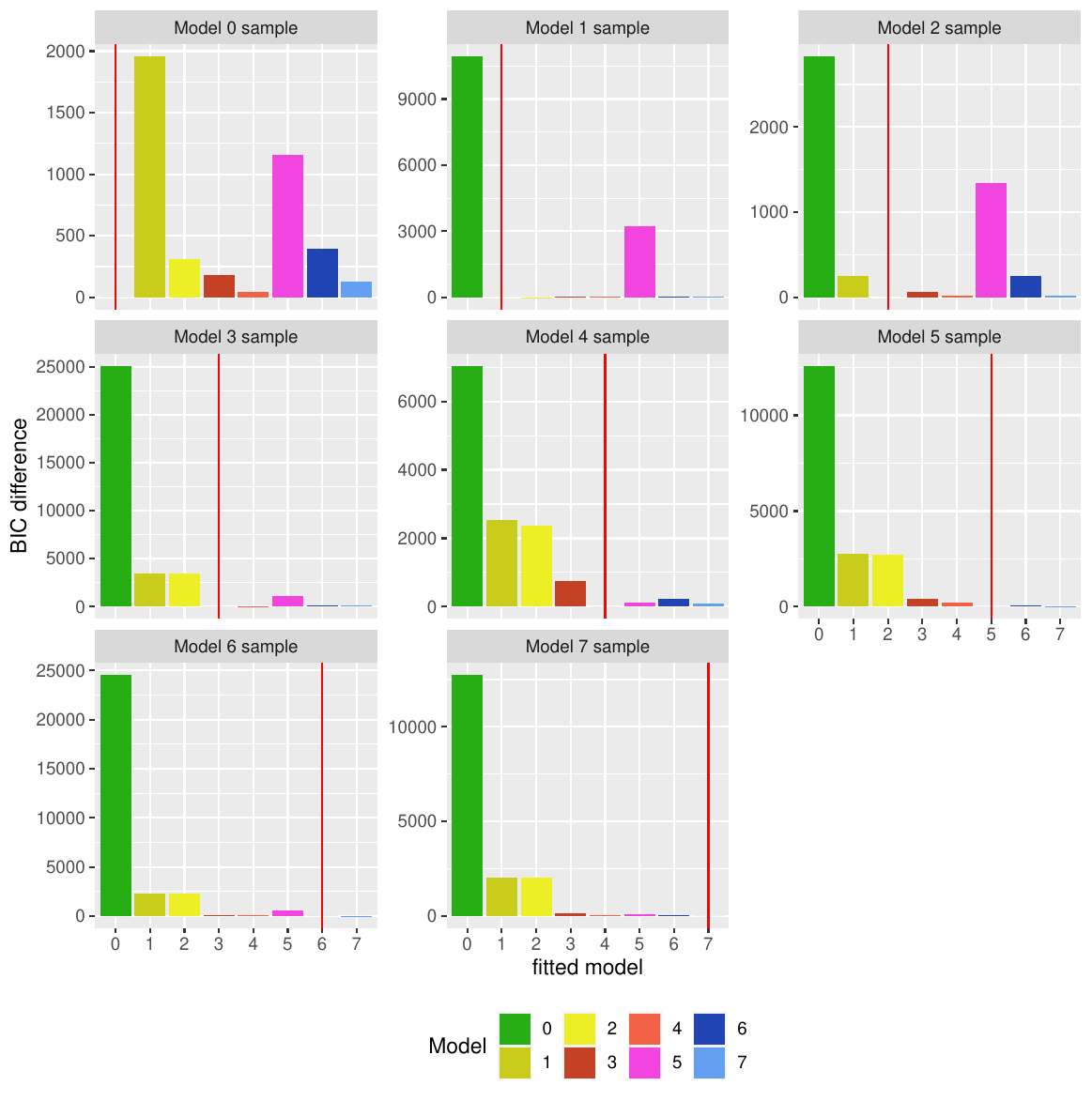}
    \caption{\label{fig:ms_artificial_BICs} 
    BIC differences in artificial random samples. The BIC difference is the difference between the BIC of the model and the BIC of the best model (the model that minimizes the BIC for the sample). The red vertical line indicates the best model according to BIC.  
    \iftoggle{Sonia}{\textcolor{red}{Remove "sample" from the subfigure titles of the form "Model X sample".}}{}
    }
\end{figure}

\begin{figure}[H]
  \centering
     \includegraphics[width = 0.85\textwidth]{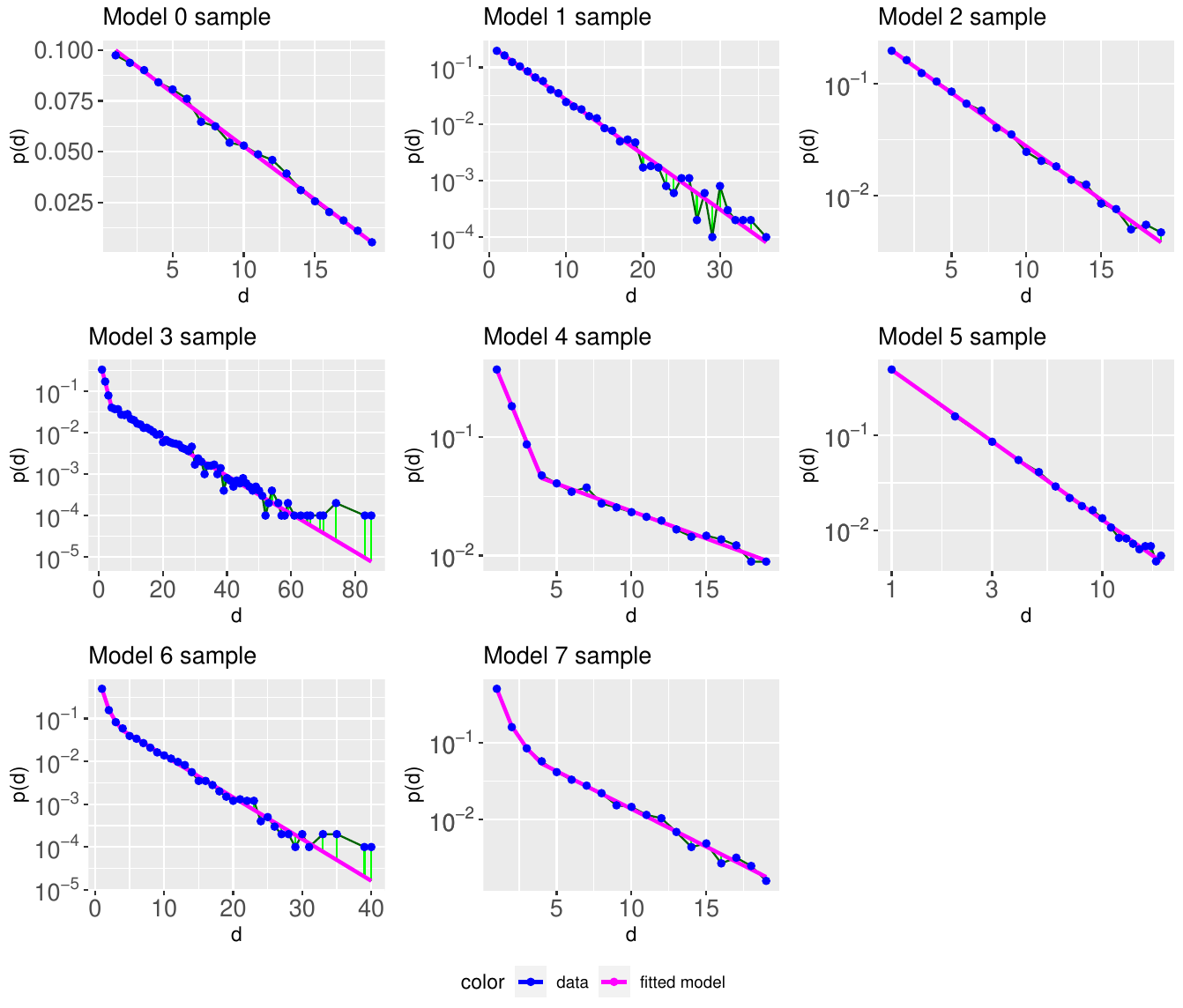}
     \caption{\label{fig:artificial_best_model}
     $p(d)$, the probability that a dependency link is formed between words at distance $d$ according to the best model for artificially generated samples. }
\end{figure}

For each model, the best model yields a good visual fit to each artificially generated sample \autoref{fig:artificial_best_model}. Indeed, the real underlying distribution is identified for every artificial random sample (\autoref{tab:artif_BIC} and \autoref{fig:ms_artificial_BICs}).  See \autoref{fig:ms_artificial_BICs} for  the magnitude of the difference in BIC score between a given model and the best model (the model that minimizes BIC). The BIC of the double-regime models is always close to the BIC of the best model. The reason resides in the greater flexibility allowed by the existence of the break-point, which is however compensated by the penalty imposed on the additional parameter by the BIC score \ref{eq:BIC}. Another concern could rise from the fitting of the random sample of Model 0, in which the BIC score of Model 4 is not much larger than that of the best model. Indeed, two geometric regimes could mimic the linearity of Model 0, but only in the case in which the second regime decays faster than the second.
\iftoggle{Sonia}{
 \textcolor{red}{Sonia/Ramon: the next sentence needs to be revised once we have a new stable version} As we discussed in section \ref{sec:Discussion}, this is virtually never the case, so that in our analysis we can exclude the risk of falsely identifying a double regime model when distances are randomly distributed.
}{}
The values of the parameters estimated by maximum likelihood for each artificially generated random sample are shown in \autoref{tab:artificial_params}.
See \autoref{tab:params_compare} for a comparison of the estimated values against the real values used to generate the data for each of the artificial samples. The error between the real values and the optimal parameters is either 0 or very small. In particular, maximum likelihood seems prone to underestimate the real value rather than the opposite. 


\begin{table}[H]
\caption{\label{tab:artif_BIC} BIC scores on artificial random samples. Each row corresponds to a random sample generated by a given model. In each row, we show first the name of the true model and then we show the AIC values of each candidate model.
\iftoggle{Sonia}{}{The true Model 0 is Model 0 that is equivalent to Model 0.1 here. The candidate Model 0 is Model 0.0.}
}
\begin{tabular}{lrrrrrrrr}
\toprule
\textbf{True model} & \textbf{Model 0} & \textbf{Model 1} & \textbf{Model 2} & \textbf{Model 3} & \textbf{Model 4} & \textbf{Model 5} & \textbf{Model 6} & \textbf{Model 7} \\ 
  \midrule
0 & 55570.65 & 57527.90 & 55881.07 & 55750.98 & 55615.79 & 56724.85 & 55963.35 & 55697.17 \\ 
  1 & 60974.42 & 50037.40 & 50040.08 & 50049.99 & 50056.13 & 53256.86 & 50054.30 & 50057.24 \\ 
  2 & 51569.46 & 48995.12 & 48739.88 & 48801.18 & 48755.09 & 50086.11 & 48993.61 & 48757.98 \\ 
  3 & 76657.57 & 54995.13 & 55004.33 & 51553.49 & 51561.32 & 52694.62 & 51681.76 & 51689.79 \\ 
  4 & 51638.78 & 47122.05 & 46967.51 & 45359.06 & 44595.90 & 44716.30 & 44818.89 & 44685.95 \\ 
  5 & 49460.37 & 39609.04 & 39602.60 & 37251.07 & 37076.27 & 36864.76 & 36937.93 & 36881.25 \\ 
  6 & 61658.20 & 39436.89 & 39446.10 & 37196.76 & 37204.83 & 37684.75 & 37133.38 & 37141.30 \\ 
  7 & 48909.08 & 38217.08 & 38217.08 & 36343.48 & 36242.39 & 36270.85 & 36239.71 & 36175.27 \\

  \bottomrule  
\end{tabular}
\end{table}

\renewcommand{\arraystretch}{1.4}


\begin{sidewaystable}[ht]
\caption{\label{tab:artificial_params} Best parameters estimated in artificial random samples by maximum likelihood. The 1st column indicates the true model while the header row indicates the candidate model. \iftoggle{Sonia}{}{Here Model 0 refers to Model 0.0.}
}
\footnotesize

\renewcommand{\arraystretch}{1.2}
\begin{tabular}{lrrrrrrrrrrrrrrrrrrrrr}
\toprule
\textbf{Model} & $\max(d) $ & $d_{max}$ & $q$ & $q$ & $d_{max}$ & $q_1$ & $q_2$ & $d^*$ & $q_1$ & $q_2$ & $d^*$ & $d_{max}$ & $d_{max}$ & $\gamma$ & $\gamma$ & $q$ & $d^*$ & $\gamma$ & $q$ & $d^*$ & $d_{max}$ \\     
\midrule
    & & \multicolumn{1}{l}{0} & 
    \multicolumn{1}{l}{1} & 
    \multicolumn{2}{l}{2} & 
    \multicolumn{3}{l}{3} & 
    \multicolumn{4}{l}{4} & 
    \multicolumn{2}{l}{5} & 
    \multicolumn{3}{l}{6} & 
    \multicolumn{4}{l}{7}\\
    \midrule
     0 &   19 & 19 & 0.142 & 0.100 &  19 & 0.088 & 0.643 &  17 & 0.071 & 0.242 &  13 & 19 & 19 & 0.522 & 0.302 & 0.343 &   13 & 0.264 & 0.206 &   11 &  19 \\ 
     1 &   36 & 36 & 0.200 & 0.200 &  36 & 0.200 & 0.543 &  34 & 0.191 & 0.203 & 5 & 36 & 36 & 1.204 & 0.274 & 0.201 &  2 & 0.278 & 0.201 &  2 &  36 \\ 
     2 &   19 & 19 & 0.210 & 0.197 &  19 & 0.197 & 0.622 &  18 & 0.199 & 0.091 &  17 & 19 & 19 & 0.985 & 0.418 & 0.221 &  4 & 0.295 & 0.198 &  2 &  19 \\ 
     3 &   85 & 85 & 0.160 & 0.160 &  85 & 0.502 & 0.101 & 4 & 0.502 & 0.101 & 4 & 85 & 85 & 1.422 & 1.373 & 0.103 &  5 & 1.373 & 0.102 &  5 &  85 \\ 
     4 &   19 & 19 & 0.228 & 0.219 &  19 & 0.549 & 0.166 & 3 & 0.503 & 0.101 & 4 & 19 & 19 & 1.242 & 1.234 & 0.644 &   18 & 1.332 & 0.097 &  6 &  19 \\ 
     5 &   19 & 19 & 0.314 & 0.313 &  19 & 0.628 & 0.201 & 3 & 0.641 & 0.180 & 3 & 19 & 19 & 1.582 & 1.578 & 0.610 &   18 & 1.588 & 0.058 &   15 &  19 \\ 
     6 &   40 & 40 & 0.317 & 0.317 &  40 & 0.623 & 0.204 & 3 & 0.624 & 0.204 & 3 & 40 & 40 & 1.718 & 1.613 & 0.201 &  4 & 1.614 & 0.201 &  4 &  40 \\ 
     7 &   19 & 19 & 0.333 & 0.332 &  19 & 0.613 & 0.221 & 3 & 0.622 & 0.206 & 3 & 19 & 19 & 1.608 & 1.541 & 0.299 &   12 & 1.610 & 0.202 &  4 &  19 \\ 

\bottomrule
\end{tabular}
\end{sidewaystable}

\renewcommand{\arraystretch}{1.2}
\begin{sidewaystable}[ht]
\caption{\label{tab:params_compare} Best estimated parameters, real parameters used to generate the artificial samples, and their difference. \iftoggle{Sonia}{}{Here Model 0 refers to Model 0.0.}
The header row indicates the true model.}
\footnotesize
\begin{tabular}{lrrrrrrrrrrrrrrrrrrrr}    
    \toprule
    &  \multicolumn{1}{l}{0} & 
    \multicolumn{1}{l}{1} & 
    \multicolumn{2}{l}{2} &
    \multicolumn{3}{l}{3} & 
    \multicolumn{4}{l}{4} & 
    \multicolumn{2}{l}{5} & 
    \multicolumn{3}{l}{6} & 
    \multicolumn{4}{l}{7}\\
    \midrule
 & $d_{max}$ & $ q$ & $ q$ & $d_{max}$ & $q_1$ & $q_2$ & $d^*$ & $q_1$ & $q_2$ & $d^*$ & $d_{max}$ & $d_{max}$ & $\gamma$ & $\gamma$ & $q$ & $d^*$ & $\gamma$ & $q$ & $d^*$ & $d_{max}$ \\   
    \midrule
estimated & 19 & 0.200 & 0.197  & 19 & 0.502 & 0.101 & 4 & 0.503 & 0.101 & 4 & 19  & 19  & 1.582  & 1.613 & 0.201 & 4 & 1.610 & 0.202 & 4 & 19 \\ 
  real    & 19 & 0.200 & 0.200  & 19 & 0.500 & 0.100 & 4 & 0.500 & 0.100 & 4 & 19  & 19  & 1.600  & 1.600 & 0.200 & 4 & 1.600 & 0.200 & 4 & 19 \\ 
  error   & 0  & 0.000 & $-0.003$ & 0  & 0.002 & 0.001 & 0 & 0.003 & 0.001 & 0 & 0 & 0 & $-0.018$ & 0.013 & 0.001 & 0 & 0.010 & 0.002 & 0 & 0 \\

    \bottomrule
\end{tabular}
\end{sidewaystable}

\section{Model selection results}
\label{sec:AppendixB}

\begin{figure}[H]
\centering
    \includegraphics[width = 0.85\textwidth]{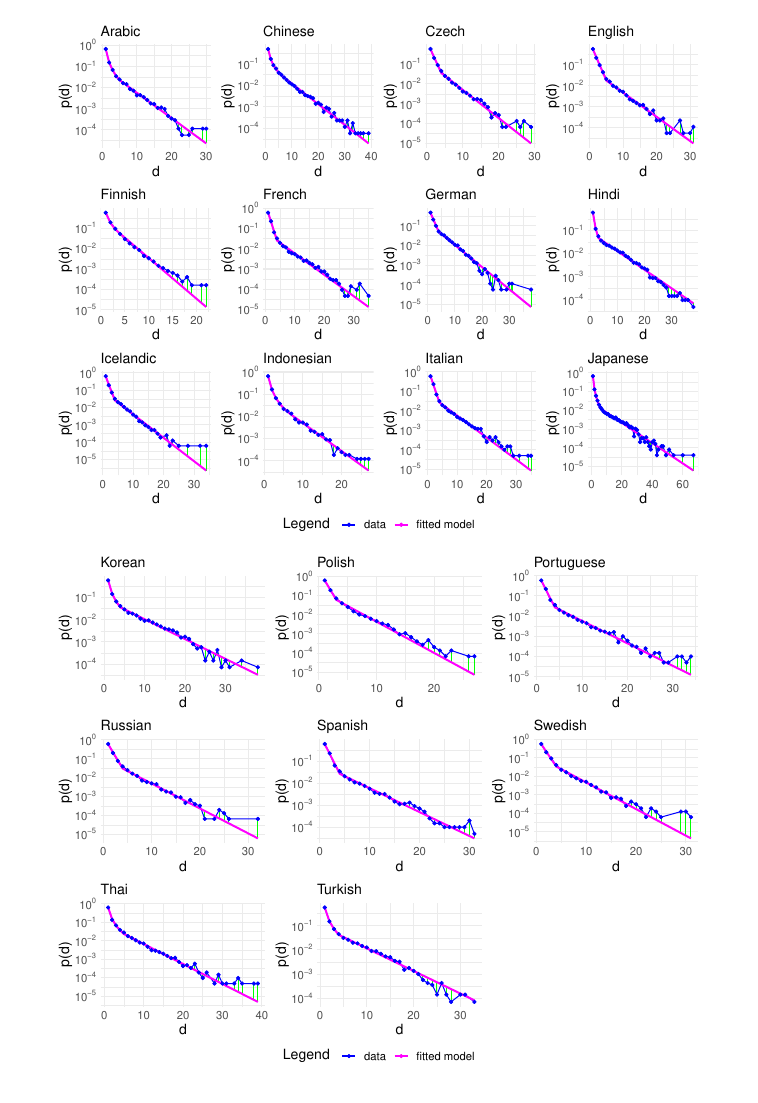}
    \caption{$p(d)$, the probability that a dependency link is formed between words at distance $d$ according to the data and the best model for every language in PSUD.} \label{fig:PSUD_best_fitted}
\end{figure}

We here report the results of model selection when sentences of any length are mixed for each language. 
See \autoref{tab:PUD_AIC_scores} and \autoref{tab:PSUD_AIC_scores} for the AIC scores for PUD and PSUD, respectively; see  \autoref{tab:PUD_AIC_diff} and \autoref{tab:PSUD_AIC_diff} for the corresponding AIC differences. The AIC difference of a model is defined as the difference of its AIC and the AIC of the best model (the model that minimizes AIC) \parencite{anderson2004model}.
The parameters estimated by maximum likelihood are shown in \autoref{tab:PUD_params} for PUD and in \autoref{tab:PSUD_params} for PSUD. Finally, see \autoref{fig:PSUD_best_fitted} for the best model fitted to the empirical distribution for languages in PSUD.

\begin{table}[H]
\caption{\label{tab:PUD_AIC_scores} AIC scores of each model in the PUD collection on sentences of mixed lengths. \iftoggle{Sonia}{\textcolor{red}{}}{Here Model 0 refers to Model 0.1.}
}
\begin{tabular}{lrrrrrrrr}
\toprule
    \textbf{Language} & \textbf{Model 0} & \textbf{Model 1} & \textbf{Model 2} & \textbf{Model 3} & \textbf{Model 4} & \textbf{Model 5} & \textbf{Model 6} & \textbf{Model 7} \\
\midrule
 Arabic & 84577 & 55188 & 55190 & 52011 & 52012 & 52264 & 51866 & 51864 \\ 
  Chinese & 86281 & 68121 & 68123 & 65826 & 65827 & 67025 & 65737 & 65738 \\ 
  Czech & 68424 & 48729 & 48731 & 47872 & 47872 & 49499 & 48212 & 48214 \\ 
  English & 85821 & 60122 & 60124 & 59402 & 59402 & 62649 & 60055 & 60056 \\ 
  Finnish & 52893 & 38423 & 38425 & 37955 & 37956 & 38921 & 37927 & 37928 \\ 
  French & 109197 & 71748 & 71750 & 69420 & 69418 & 72291 & 70944 & 70946 \\ 
  German & 86626 & 68510 & 68512 & 66699 & 66700 & 68821 & 66955 & 66957 \\ 
  Hindi & 107388 & 83075 & 83077 & 75832 & 75828 & 76676 & 75788 & 75782 \\ 
  Icelandic & 73752 & 50242 & 50244 & 49411 & 49413 & 51252 & 49716 & 49718 \\ 
  Indonesian & 76351 & 50676 & 50678 & 48875 & 48876 & 49596 & 48916 & 48917 \\ 
  Italian & 104223 & 68313 & 68315 & 66370 & 66369 & 69289 & 67786 & 67788 \\ 
  Japanese & 135512 & 93746 & 93748 & 85222 & 85221 & 87112 & 86524 & 86525 \\ 
  Korean & 64173 & 50365 & 50367 & 45474 & 45472 & 45647 & 45337 & 45332 \\ 
  Polish & 66255 & 45103 & 45105 & 43851 & 43852 & 44719 & 43956 & 43958 \\ 
  Portuguese & 100042 & 67213 & 67215 & 65361 & 65361 & 68010 & 66557 & 66559 \\ 
  Russian & 70474 & 47879 & 47881 & 46750 & 46751 & 48291 & 47201 & 47203 \\ 
  Spanish & 101194 & 67353 & 67355 & 65377 & 65376 & 67934 & 66641 & 66643 \\ 
  Swedish & 75639 & 53807 & 53809 & 53135 & 53136 & 55623 & 53633 & 53635 \\ 
  Thai & 108081 & 69553 & 69555 & 65717 & 65718 & 66242 & 65519 & 65521 \\ 
  Turkish & 62864 & 51439 & 51441 & 47362 & 47358 & 47697 & 47250 & 47245 \\ 

\bottomrule
\end{tabular}
\end{table}

\begin{table}[H]
\caption{\label{tab:PUD_AIC_diff} AIC differences of each model in the PUD collection on sentences of mixed lengths. \iftoggle{Sonia}{\textcolor{red}{}}{Here Model 0 refers to Model 0.1.}
}
\begin{tabular}{lrrrrrrrr}
\toprule
    \textbf{Language} & \textbf{Model 0} & \textbf{Model 1} & \textbf{Model 2} & \textbf{Model 3} & \textbf{Model 4} & \textbf{Model 5} & \textbf{Model 6} & \textbf{Model 7} \\
\midrule
 Arabic & 32712.96 & 3323.95 & 3325.95 & 146.57 & 147.28 & 399.60 & 1.44 & 0.00 \\ 
  Chinese & 20544.43 & 2383.68 & 2385.68 & 88.77 & 90.07 & 1288.62 & 0.00 & 0.72 \\ 
  Czech & 20552.34 & 857.52 & 859.51 & 0.00 & 0.70 & 1627.44 & 340.07 & 341.97 \\ 
  English & 26419.25 & 720.65 & 722.64 & 0.00 & 0.63 & 3247.77 & 652.91 & 654.90 \\ 
  Finnish & 14965.50 & 496.10 & 498.00 & 27.96 & 29.28 & 993.94 & 0.00 & 1.02 \\ 
  French & 39779.35 & 2329.86 & 2331.86 & 1.94 & 0.00 & 2873.31 & 1526.42 & 1528.33 \\ 
  German & 19927.31 & 1811.19 & 1813.19 & 0.00 & 1.49 & 2122.61 & 256.63 & 258.34 \\ 
  Hindi & 31605.72 & 7292.04 & 7294.03 & 49.53 & 45.49 & 893.65 & 5.61 & 0.00 \\ 
  Icelandic & 24340.78 & 831.13 & 833.13 & 0.00 & 1.94 & 1841.66 & 305.52 & 307.52 \\ 
  Indonesian & 27476.04 & 1800.38 & 1802.38 & 0.00 & 1.08 & 721.10 & 41.17 & 41.86 \\ 
  Italian & 37854.79 & 1944.06 & 1946.06 & 1.10 & 0.00 & 2920.54 & 1417.74 & 1419.67 \\ 
  Japanese & 50290.71 & 8524.56 & 8526.56 & 0.58 & 0.00 & 1890.84 & 1302.96 & 1303.51 \\ 
  Korean & 18840.12 & 5032.98 & 5034.98 & 141.29 & 139.91 & 314.13 & 4.31 & 0.00 \\ 
  Polish & 22403.20 & 1251.25 & 1253.25 & 0.00 & 0.73 & 867.34 & 104.69 & 106.27 \\ 
  Portuguese & 34681.44 & 1852.22 & 1854.22 & 0.00 & 0.34 & 2649.16 & 1196.46 & 1198.33 \\ 
  Russian & 23723.74 & 1129.28 & 1131.28 & 0.00 & 1.37 & 1540.80 & 451.36 & 453.32 \\ 
  Spanish & 35817.93 & 1976.75 & 1978.74 & 1.07 & 0.00 & 2557.41 & 1264.87 & 1266.59 \\ 
  Swedish & 22503.61 & 671.54 & 673.54 & 0.00 & 0.88 & 2487.59 & 497.78 & 499.75 \\ 
  Thai & 42561.20 & 4033.29 & 4035.29 & 197.05 & 198.82 & 722.50 & 0.00 & 1.23 \\ 
  Turkish & 15618.79 & 4193.97 & 4195.96 & 117.05 & 112.74 & 452.16 & 5.51 & 0.00 \\
\bottomrule
\end{tabular}
\end{table}


 
\begin{table}[H]
\caption{\label{tab:PSUD_AIC_scores} AIC scores of each model in the PSUD collection on sentences of mixed lengths. \iftoggle{Sonia}{\textcolor{red}{}}{Here Model 0 refers to Model 0.1.}
}
\begin{tabular}{lrrrrrrrr}    
\toprule
    \textbf{Language} & \textbf{Model 0} & \textbf{Model 1} & \textbf{Model 2} & \textbf{Model 3} & \textbf{Model 4} & \textbf{Model 5} & \textbf{Model 6} & \textbf{Model 7} \\
\midrule
 Arabic & 83964 & 49718 & 49720 & 45461 & 45461 & 45444 & 45249 & 45248 \\ 
  Chinese & 86004 & 66658 & 66660 & 63187 & 63188 & 63852 & 63043 & 63043 \\ 
  Czech & 67743 & 43660 & 43662 & 42048 & 42049 & 42711 & 42164 & 42166 \\ 
  English & 84875 & 51868 & 51870 & 49860 & 49860 & 50895 & 50293 & 50295 \\ 
  Finnish & 52389 & 35247 & 35249 & 34450 & 34451 & 35057 & 34434 & 34436 \\ 
  French & 108313 & 62309 & 62311 & 57458 & 57458 & 58294 & 58037 & 58036 \\ 
  German & 85580 & 64289 & 64291 & 62110 & 62111 & 63642 & 62229 & 62230 \\ 
  Hindi & 106846 & 79001 & 79003 & 68777 & 68760 & 69540 & 68495 & 68483 \\ 
  Icelandic & 72807 & 42153 & 42155 & 39927 & 39929 & 40396 & 40000 & 40002 \\ 
  Indonesian & 75765 & 45212 & 45214 & 41927 & 41928 & 42005 & 41809 & 41808 \\ 
  Italian & 103370 & 59445 & 59447 & 55354 & 55354 & 56373 & 56039 & 56040 \\ 
  Japanese & 135293 & 88560 & 88562 & 72667 & 72667 & 72316 & 71831 & 71831 \\ 
  Korean & 64065 & 49797 & 49799 & 44509 & 44508 & 44683 & 44366 & 44364 \\ 
  Polish & 65708 & 40765 & 40767 & 38674 & 38676 & 39020 & 38697 & 38698 \\ 
  Portuguese & 99155 & 58440 & 58442 & 54381 & 54381 & 55150 & 54846 & 54846 \\ 
  Russian & 69999 & 43597 & 43599 & 41455 & 41457 & 41963 & 41541 & 41543 \\ 
  Spanish & 100350 & 58907 & 58909 & 54617 & 54615 & 55352 & 55105 & 55103 \\ 
  Swedish & 74683 & 46288 & 46290 & 44584 & 44586 & 45399 & 44815 & 44817 \\ 
  Thai & 107549 & 63835 & 63837 & 57879 & 57881 & 57879 & 57564 & 57565 \\ 
  Turkish & 62784 & 50771 & 50773 & 46448 & 46442 & 46728 & 46325 & 46318 \\ 
\bottomrule    
\end{tabular}
\end{table}

\begin{table}[H]
\caption{\label{tab:PSUD_AIC_diff} AIC differences of each model in the PSUD collection on sentences of mixed lengths. \iftoggle{Sonia}{\textcolor{red}{}}{Here Model 0 refers to Model 0.1.}
}
\begin{tabular}{lrrrrrrrr}    
\toprule
    \textbf{Language} & \textbf{Model 0} & \textbf{Model 1} & \textbf{Model 2} & \textbf{Model 3} & \textbf{Model 4} & \textbf{Model 5} & \textbf{Model 6} & \textbf{Model 7} \\
\midrule
 Arabic & 38715.46 & 4469.92 & 4471.92 & 212.61 & 213.36 & 196.35 & 0.49 & 0.00 \\ 
  Chinese & 22961.93 & 3615.17 & 3617.16 & 144.40 & 145.21 & 809.83 & 0.67 & 0.00 \\ 
  Czech & 25695.55 & 1612.08 & 1614.08 & 0.00 & 1.23 & 663.54 & 116.10 & 117.76 \\ 
  English & 35015.25 & 2008.17 & 2010.17 & 0.09 & 0.00 & 1034.96 & 433.57 & 435.32 \\ 
  Finnish & 17954.96 & 812.85 & 814.84 & 15.62 & 16.93 & 622.93 & 0.00 & 1.41 \\ 
  French & 50855.16 & 4851.44 & 4853.44 & 0.05 & 0.00 & 836.35 & 579.48 & 578.26 \\ 
  German & 23469.70 & 2179.07 & 2181.07 & 0.00 & 1.12 & 1532.51 & 118.75 & 120.29 \\ 
  Hindi & 38363.24 & 10518.04 & 10520.03 & 294.32 & 277.30 & 1056.95 & 11.70 & 0.00 \\ 
  Icelandic & 32879.81 & 2225.42 & 2227.42 & 0.00 & 1.79 & 468.28 & 72.30 & 74.23 \\ 
  Indonesian & 33956.92 & 3404.16 & 3406.16 & 119.39 & 119.95 & 196.85 & 1.41 & 0.00 \\ 
  Italian & 48016.12 & 4091.05 & 4093.05 & 0.00 & 0.83 & 1019.71 & 685.68 & 686.27 \\ 
  Japanese & 63462.28 & 16729.23 & 16731.23 & 836.09 & 836.29 & 485.06 & 0.28 & 0.00 \\ 
  Korean & 19701.40 & 5432.90 & 5434.90 & 145.71 & 144.65 & 319.57 & 1.91 & 0.00 \\ 
  Polish & 27034.25 & 2090.57 & 2092.57 & 0.00 & 1.52 & 346.04 & 23.00 & 23.88 \\ 
  Portuguese & 44774.69 & 4059.49 & 4061.49 & 0.00 & 0.29 & 769.23 & 465.71 & 465.70 \\ 
  Russian & 28543.92 & 2141.67 & 2143.67 & 0.00 & 1.47 & 508.13 & 86.14 & 87.86 \\ 
  Spanish & 45735.15 & 4292.48 & 4294.48 & 2.28 & 0.00 & 736.75 & 490.45 & 487.74 \\ 
  Swedish & 30099.05 & 1703.55 & 1705.55 & 0.00 & 1.64 & 815.39 & 231.31 & 233.16 \\ 
  Thai & 49985.45 & 6271.66 & 6273.66 & 315.48 & 317.16 & 315.43 & 0.00 & 1.29 \\ 
  Turkish & 16466.09 & 4453.49 & 4455.47 & 129.96 & 124.00 & 410.41 & 7.45 & 0.00 \\ 
\bottomrule    
\end{tabular}
\end{table}



\begin{sidewaystable}[ht]
\caption{\label{tab:PUD_params} Best parameters estimated by maximum likelihood on sentences of mixed lengths in the PUD collection.}
\footnotesize
\begin{tabular}{lrrrrrrrrrrrrrrrrrrrrr}
\toprule
    & &  & \multicolumn{1}{c}{1}
    & \multicolumn{2}{c}{2}
    & \multicolumn{3}{c}{3}
    & \multicolumn{4}{c}{4}
    & \multicolumn{2}{c}{5}
    & \multicolumn{3}{c}{6}
    & \multicolumn{4}{c}{7}\\
\midrule    
\textbf{Language} & $\max(n)$ & $\max(d)$ & $q$ & $q$ & $d_{max}$ & $q_1$ & $q_2$ & $d^*$ & $q_1$ & $q_2$ & $d^*$ & $d_{max}$ & $d_{max}$ & $\gamma$ & $\gamma$ & $q$ & $d^*$ & $\gamma$ & $q$ & $d^*$ & $d_{max}$ \\ 
\midrule
Arabic &   50 &   30 & 0.434 & 0.434 &   30 & 0.668 & 0.269 &    3 & 0.668 & 0.269 &    3 &   30 &   30 & 1.973 & 1.840 & 0.240 &    7 & 1.842 & 0.238 &    7 &   30 \\ 
  Czech &   44 &   29 & 0.418 & 0.418 &   29 & 0.499 & 0.270 &    5 & 0.500 & 0.269 &    5 &   29 &   29 & 1.837 & 1.385 & 0.347 &    3 & 1.385 & 0.347 &    3 &   29 \\ 
  German &   50 &   42 & 0.322 & 0.322 &   42 & 0.485 & 0.222 &    4 & 0.485 & 0.222 &    4 &   42 &   42 & 1.675 & 1.351 & 0.234 &    5 & 1.351 & 0.234 &    5 &   42 \\ 
  English &   56 &   31 & 0.395 & 0.395 &   31 & 0.453 & 0.256 &    6 & 0.454 & 0.255 &    6 &   31 &   31 & 1.759 & 0.930 & 0.380 &    2 & 0.930 & 0.380 &    2 &   31 \\ 
  Finnish &   39 &   21 & 0.446 & 0.446 &   21 & 0.639 & 0.388 &    2 & 0.562 & 0.360 &    3 &   21 &   21 & 1.855 & 1.440 & 0.374 &    3 & 1.440 & 0.374 &    3 &   21 \\ 
  French &   54 &   36 & 0.396 & 0.396 &   36 & 0.491 & 0.197 &    6 & 0.492 & 0.196 &    6 &   36 &   36 & 1.826 & 1.462 & 0.296 &    4 & 1.462 & 0.296 &    4 &   36 \\ 
  Hindi &   58 &   42 & 0.303 & 0.303 &   42 & 0.671 & 0.175 &    3 & 0.672 & 0.174 &    3 &   42 &   42 & 1.735 & 1.798 & 0.170 &    4 & 1.800 & 0.169 &    4 &   42 \\ 
  Indonesian &   47 &   27 & 0.442 & 0.442 &   27 & 0.629 & 0.305 &    3 & 0.630 & 0.304 &    3 &   27 &   27 & 1.935 & 1.713 & 0.295 &    5 & 1.714 & 0.294 &    5 &   27 \\ 
  Icelandic &   52 &   34 & 0.432 & 0.432 &   34 & 0.531 & 0.309 &    4 & 0.531 & 0.309 &    4 &   34 &   34 & 1.888 & 1.412 & 0.359 &    3 & 1.412 & 0.359 &    3 &   34 \\ 
  Italian &   60 &   35 & 0.403 & 0.403 &   35 & 0.490 & 0.207 &    6 & 0.491 & 0.206 &    6 &   35 &   35 & 1.830 & 1.446 & 0.307 &    4 & 1.446 & 0.307 &    4 &   35 \\ 
  Japanese &   70 &   65 & 0.337 & 0.337 &   65 & 0.521 & 0.119 &    6 & 0.522 & 0.118 &    6 &   65 &   65 & 1.849 & 1.754 & 0.130 &   13 & 1.755 & 0.130 &   13 &   65 \\ 
  Korean &   43 &   37 & 0.364 & 0.364 &   37 & 0.700 & 0.197 &    3 & 0.701 & 0.197 &    3 &   37 &   37 & 1.886 & 1.886 & 0.180 &    5 & 1.888 & 0.179 &    5 &   37 \\ 
  Polish &   39 &   27 & 0.449 & 0.449 &   27 & 0.569 & 0.288 &    4 & 0.569 & 0.287 &    4 &   27 &   27 & 1.936 & 1.653 & 0.324 &    4 & 1.653 & 0.324 &    4 &   27 \\ 
  Portuguese &   58 &   34 & 0.396 & 0.396 &   34 & 0.504 & 0.234 &    5 & 0.504 & 0.233 &    5 &   34 &   34 & 1.812 & 1.436 & 0.301 &    4 & 1.436 & 0.301 &    4 &   34 \\ 
  Russian &   47 &   32 & 0.440 & 0.440 &   32 & 0.529 & 0.261 &    5 & 0.529 & 0.260 &    5 &   32 &   32 & 1.916 & 1.564 & 0.332 &    4 & 1.564 & 0.332 &    4 &   32 \\ 
  Spanish &   58 &   32 & 0.399 & 0.399 &   32 & 0.510 & 0.231 &    5 & 0.510 & 0.230 &    5 &   32 &   32 & 1.816 & 1.460 & 0.300 &    4 & 1.460 & 0.300 &    4 &   32 \\ 
  Swedish &   49 &   31 & 0.404 & 0.404 &   31 & 0.462 & 0.257 &    6 & 0.462 & 0.257 &    6 &   31 &   31 & 1.791 & 1.214 & 0.358 &    3 & 1.214 & 0.358 &    3 &   31 \\ 
  Thai &   63 &   38 & 0.409 & 0.409 &   38 & 0.653 & 0.258 &    3 & 0.653 & 0.258 &    3 &   38 &   38 & 1.933 & 1.770 & 0.230 &    7 & 1.770 & 0.230 &    7 &   38 \\ 
  Turkish &   37 &   34 & 0.343 & 0.343 &   34 & 0.670 & 0.201 &    3 & 0.671 & 0.200 &    3 &   34 &   34 & 1.797 & 1.812 & 0.195 &    4 & 1.815 & 0.194 &    4 &   34 \\ 
  Chinese &   49 &   39 & 0.323 & 0.323 &   39 & 0.569 & 0.233 &    3 & 0.569 & 0.232 &    3 &   39 &   39 & 1.694 & 1.439 & 0.219 &    6 & 1.440 & 0.219 &    6 &   39 \\ 

\bottomrule
\end{tabular}
\end{sidewaystable}

\begin{sidewaystable}[ht]
\caption{\label{tab:PSUD_params} Best parameters estimated by maximum likelihood on sentences of mixed lengths in the PSUD collection.}
\footnotesize
\begin{tabular}{lrrrrrrrrrrrrrrrrrrrrr}
\toprule
    & &  & \multicolumn{1}{c}{1}
    & \multicolumn{2}{c}{2}
    & \multicolumn{3}{c}{3}
    & \multicolumn{4}{c}{4}
    & \multicolumn{2}{c}{5}
    & \multicolumn{3}{c}{6}
    & \multicolumn{4}{c}{7}\\
\midrule
\textbf{Language} & $\max(d)$ & $\max(n)$ & $q$ & $q$ & $d_{max}$ & $q_1$ & $q_2$ & $d^*$ & $q_1$ & $q_2$ & $d^*$ & $d_{max}$ & $d_{max}$ & $\gamma$ & $\gamma$ & $q$ & $d^*$ & $\gamma$ & $q$ & $d^*$ & $d_{max}$ \\
\midrule
Arabic &   50 &   30 & 0.488 & 0.488 &   30 & 0.727 & 0.263 &    3 & 0.727 & 0.263 &    3 &   30 &   30 & 2.150 & 2.101 & 0.241 &    5 & 2.102 & 0.241 &    5 &   30 \\ 
  Czech &   44 &   29 & 0.475 & 0.475 &   29 & 0.602 & 0.279 &    4 & 0.602 & 0.279 &    4 &   29 &   29 & 2.027 & 1.814 & 0.306 &    5 & 1.814 & 0.306 &    5 &   29 \\ 
  German &   50 &   38 & 0.355 & 0.355 &   38 & 0.523 & 0.228 &    4 & 0.523 & 0.228 &    4 &   38 &   38 & 1.758 & 1.489 & 0.244 &    5 & 1.489 & 0.244 &    5 &   38 \\ 
  English &   56 &   31 & 0.472 & 0.472 &   31 & 0.575 & 0.234 &    5 & 0.575 & 0.233 &    5 &   31 &   31 & 2.025 & 1.810 & 0.299 &    5 & 1.810 & 0.299 &    5 &   31 \\ 
  Finnish &   39 &   22 & 0.490 & 0.490 &   22 & 0.625 & 0.362 &    3 & 0.625 & 0.362 &    3 &   22 &   22 & 2.007 & 1.715 & 0.369 &    4 & 1.715 & 0.369 &    4 &   22 \\ 
  French &   54 &   35 & 0.470 & 0.470 &   35 & 0.652 & 0.216 &    4 & 0.652 & 0.216 &    4 &   35 &   35 & 2.090 & 2.007 & 0.206 &    9 & 2.008 & 0.204 &    9 &   35 \\ 
  Hindi &   58 &   38 & 0.329 & 0.329 &   38 & 0.729 & 0.162 &    3 & 0.730 & 0.161 &    3 &   38 &   38 & 1.843 & 2.298 & 0.171 &    3 & 2.302 & 0.170 &    3 &   38 \\ 
  Indonesian &   47 &   27 & 0.500 & 0.500 &   27 & 0.717 & 0.283 &    3 & 0.717 & 0.283 &    3 &   27 &   27 & 2.150 & 2.055 & 0.254 &    7 & 2.056 & 0.251 &    7 &   27 \\ 
  Icelandic &   52 &   34 & 0.521 & 0.521 &   34 & 0.653 & 0.266 &    4 & 0.653 & 0.266 &    4 &   34 &   34 & 2.184 & 2.029 & 0.295 &    6 & 2.030 & 0.295 &    6 &   34 \\ 
  Italian &   60 &   35 & 0.475 & 0.475 &   35 & 0.645 & 0.229 &    4 & 0.645 & 0.228 &    4 &   35 &   35 & 2.087 & 1.981 & 0.226 &    8 & 1.982 & 0.225 &    8 &   35 \\ 
  Japanese &   70 &   67 & 0.367 & 0.367 &   67 & 0.699 & 0.121 &    4 & 0.699 & 0.121 &    4 &   67 &   67 & 2.060 & 2.218 & 0.117 &    6 & 2.218 & 0.117 &    6 &   67 \\ 
  Korean &   43 &   38 & 0.370 & 0.370 &   38 & 0.713 & 0.193 &    3 & 0.713 & 0.193 &    3 &   38 &   38 & 1.915 & 2.016 & 0.187 &    4 & 2.018 & 0.186 &    4 &   38 \\ 
  Polish &   39 &   27 & 0.501 & 0.501 &   27 & 0.688 & 0.313 &    3 & 0.688 & 0.313 &    3 &   27 &   27 & 2.112 & 1.970 & 0.287 &    6 & 1.971 & 0.286 &    6 &   27 \\ 
  Portuguese &   58 &   34 & 0.469 & 0.469 &   34 & 0.642 & 0.226 &    4 & 0.642 & 0.225 &    4 &   34 &   34 & 2.074 & 1.971 & 0.224 &    8 & 1.972 & 0.223 &    8 &   34 \\ 
  Russian &   47 &   32 & 0.489 & 0.489 &   32 & 0.629 & 0.262 &    4 & 0.629 & 0.262 &    4 &   32 &   32 & 2.092 & 1.930 & 0.282 &    6 & 1.930 & 0.282 &    6 &   32 \\ 
  Spanish &   58 &   31 & 0.469 & 0.469 &   31 & 0.646 & 0.222 &    4 & 0.647 & 0.221 &    4 &   31 &   31 & 2.071 & 1.981 & 0.221 &    8 & 1.982 & 0.219 &    8 &   31 \\ 
  Swedish &   49 &   31 & 0.482 & 0.482 &   31 & 0.607 & 0.283 &    4 & 0.607 & 0.283 &    4 &   31 &   31 & 2.050 & 1.832 & 0.309 &    5 & 1.832 & 0.309 &    5 &   31 \\ 
  Thai &   63 &   39 & 0.454 & 0.454 &   39 & 0.723 & 0.240 &    3 & 0.723 & 0.240 &    3 &   39 &   39 & 2.100 & 2.053 & 0.221 &    5 & 2.053 & 0.221 &    5 &   39 \\ 
  Turkish &   37 &   33 & 0.350 & 0.350 &   33 & 0.680 & 0.199 &    3 & 0.681 & 0.198 &    3 &   33 &   33 & 1.818 & 1.859 & 0.194 &    4 & 1.863 & 0.193 &    4 &   33 \\ 
  Chinese &   49 &   39 & 0.335 & 0.335 &   39 & 0.619 & 0.219 &    3 & 0.619 & 0.219 &    3 &   39 &   39 & 1.755 & 1.574 & 0.200 &    7 & 1.574 & 0.199 &    7 &   39 \\ 

\bottomrule
\end{tabular}
\end{sidewaystable}

\section{The distribution of dependency distances for characteristic sentence lengths.}
\label{sec:AppendixC}

\begin{figure}[H]
\centering
    \includegraphics[width = 0.85\textwidth]{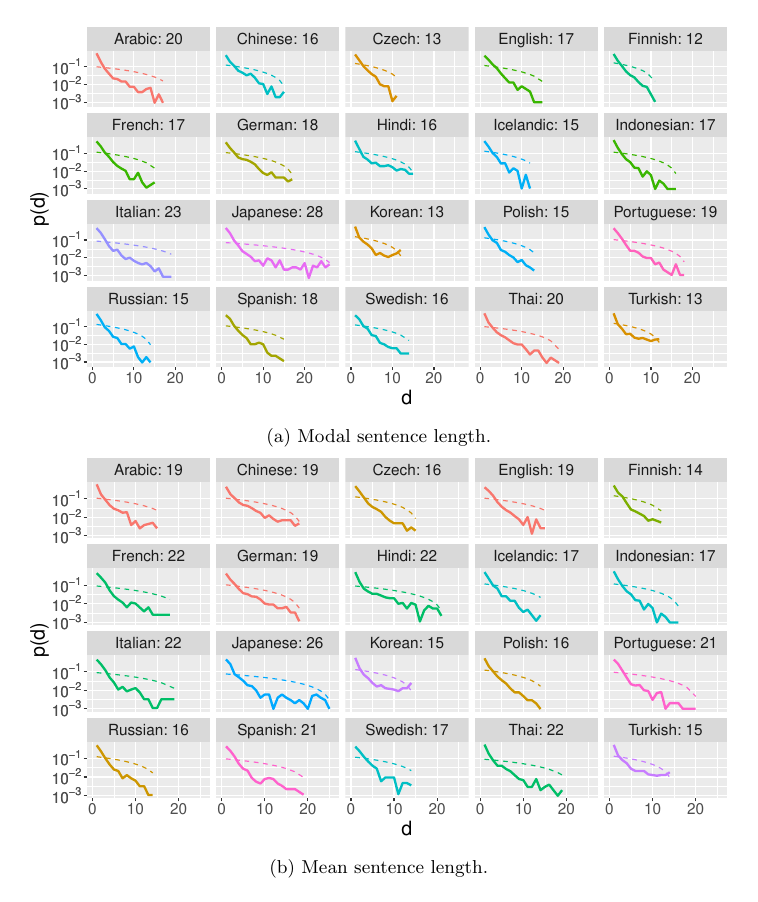}
    \caption{\label{fig:probability_mean_modal_sentence_length_PUD} $p(d)$, the probability that linked words are at distance $d$ in sentences of modal (a) and mean (b) length for each language in PUD. Mode and mean are shown next to the respective language label. The dashed line shows the probability according to Model 0 (\ref{eq:model0}). Points where $p(d)=0$ are not shown.}
\end{figure}

\begin{figure}[H]
\centering
    \includegraphics[width = 0.85\textwidth]{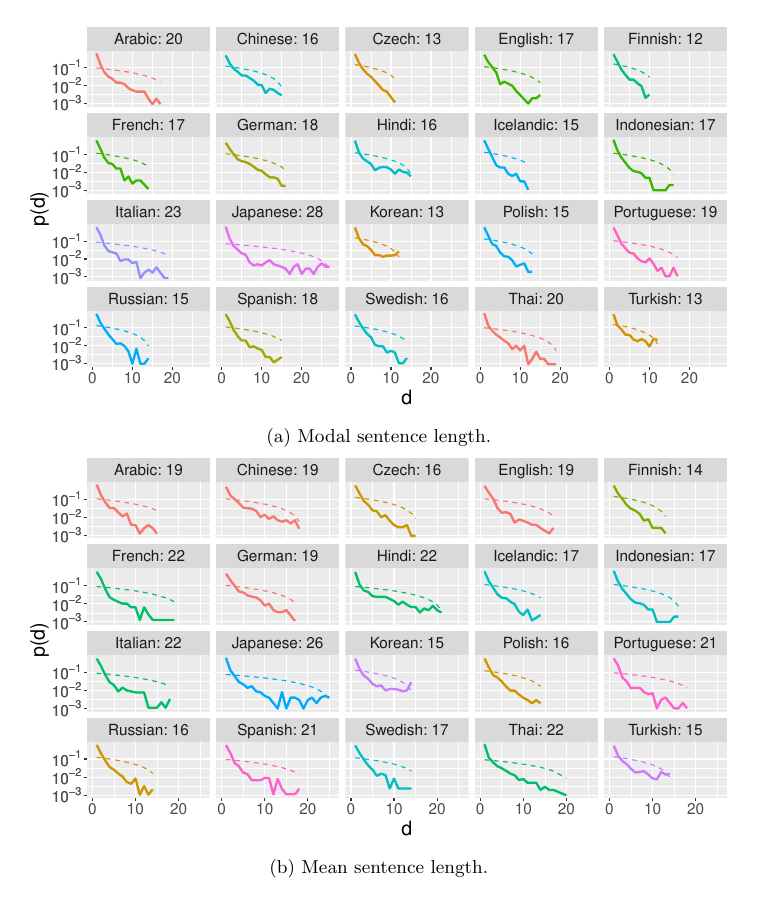}
    \caption{\label{fig:probability_mean_modal_sentence_length_PSUD} $p(d)$, the probability that linked words are at distance $d$ in sentences of modal (a) and mean (b) length for each language in PSUD. The format is the same as in  \autoref{fig:probability_mean_modal_sentence_length_PUD}.}
\end{figure}

See  \autoref{fig:probability_mean_modal_sentence_length_PUD} (a-b) for the distributions in PUD, for modal and mean sentence length respectively; see \autoref{fig:probability_mean_modal_sentence_length_PSUD} (a-b) for PSUD.
As mean sentence length, we use the results of rounding the actual mean sentence length to the nearest integer. \iftoggle{Sonia}{
\textcolor{red}{Sonia: is the statement correct?}
}{}

\end{document}